\documentclass[journal]{IEEEtran}
\usepackage{url}

\usepackage[utf8]{inputenc}
\usepackage{cite}
\usepackage{graphicx}
\usepackage{subfig}
\usepackage{float}
\usepackage{enumitem}
\usepackage{bm}
\usepackage{amsmath,amssymb,amsfonts}
\usepackage{amsthm}
\usepackage{tabulary}
\usepackage{multirow}
\usepackage{booktabs}
\usepackage{chngpage}
\usepackage{color}
\usepackage{xcolor}
\usepackage{amsmath}
\allowdisplaybreaks[4]

\newcommand{\wang}[1]{{\color{black} #1}}

\newcommand{\ywunew}[1]{{\color{black} #1}}

\begin{document}
%
\title{Efficient Visual Recognition with Deep Neural Networks: A Survey on Recent Advances\\and New Directions}
%
%
%


\author{Yang Wu, Dingheng Wang, Xiaotong Lu, Fan Yang, Guoqi Li\dag, Weisheng Dong, Jianbo Shi
\thanks{Y. Wu is with Institute for Research Initiatives, Nara Institute of Science and Technology, Takayama-cho, Ikoma, Nara 630-0192, Japan.}
\thanks{D. Wang is with School of Automation Science and Engineering, Faculty of Electronic and Information Engineering, Xi'an Jiaotong University, Xi'an, Shaanxi 710049, China.}
\thanks{X. Lu and W. Dong are with School of Artificial Intelligence, Xidian University, China.}
\thanks{F. Yang is with Division of Information Science, Nara Institute of Science and Technology, Japan.}
\thanks{G. Li is with Department of Precision Instrumentation, Center for Brain Inspired Computing Research and  Beijing Innovation Center for Future Chip, Tsinghua University, Beijing 100084, China.}
\thanks{J. Shi is with Department of Computer and Information Science, University of Pennsylvania, United States.}
\thanks{\dag: Corresponding author, \protect\url{liguoqi@mail.tsinghua.edu.cn}.}}

\maketitle

\begin{abstract}
Visual recognition is currently one of the most important and active research areas in computer vision, pattern recognition, and even the general field of artificial intelligence. It has great fundamental importance and strong industrial needs. Deep neural networks (DNNs) have largely boosted their performances on many concrete tasks, with the help of large amounts of training data and new powerful computation resources. Though recognition accuracy is usually the first concern for new progresses, efficiency is actually rather important and sometimes critical for both academic research and industrial applications. Moreover, insightful views on the opportunities and challenges of efficiency are also highly required for the entire community. While general surveys on the efficiency issue of DNNs have been done from various perspectives, as far as we are aware, \wang{scarcely any of them} focused on visual recognition systematically, and thus it is unclear which progresses are applicable to it and what else should be concerned. In this paper, we present the review of the recent advances with our suggestions on the new possible directions towards improving the efficiency of DNN-related visual recognition approaches. We investigate not only from  the model but also the data point of view (which is not the case in existing surveys), and focus on three most studied data types (images, videos and points). This paper attempts to  provide  a systematic summary via a comprehensive survey which can serve as a valuable reference and inspire both researchers and practitioners who work on visual recognition problems.
\end{abstract}


%
\IEEEpeerreviewmaketitle

\section{Introduction}
\label{intro}

Deep neural networks (DNNs) have achieved great successes on many visual recognition tasks. They have largely improved the performances of long-lasting problems such as handwritten digit recognition \cite{Lecun_1998_MNIST_PIEEE}, face recognition \cite{Hinton_2006_Science}, image categorization \cite{Krizhevsky_2012_AlexNet_NIPS}, etc. They are also enabling exploring new boundaries including the studies on image and video captioning \cite{Lin_2014_COCO_ECCV,Song_2019_MS-RNNs_TNNLS,Gao_2019_HLSTMs_PAMI}, body pose estimation \cite{Wei_2016_CPM_CVPR}, and many others. However, such successes are generally conditioned on huge amounts of high-quality hand labelled training data and the recently greatly advanced computational resources. Obviously, these two conditions are usually too expensive to be satisfied in most applications which are cost-sensitive. Even for the cases that \wang{people} do have enough high-quality training data thanks to the massive efforts of many annotators, it is usually a great challenge to figure out how to train an effective model with limited resources and within acceptable time. Assuming somehow the model can be properly trained (no matter how much efforts it costs), it is still not easy to have the model properly deployed for real applications at end users' side, as the run-time inference has to fit the available or affordable resources, and the running speed has to meet the actual needs which can be real-time or even more than that. Therefore, besides accuracy which is usually the biggest concern in the academia, efficiency is another important issue and in most cases an indispensable demand for real applications.

Though most of the works on using DNNs for visual recognition tasks are focusing on accuracy, there are still a lot of encouraging progresses on the efficiency side, especially in the recent few years. In the past two years, many survey papers have been published on efficiency issue for DNNs, as detailed in the following subsection \ref{subsec:related_surveys}. However, none of them pays a major attention to visual recognition tasks, especially missing the coverage of special efforts for efficiently dealing with visual data, which has its own properties. In practice, efficient visual recognition has to be a systematic solution which takes into account of not only compact/compressed networks and hardware acceleration, but also proper handling of visual data, which may be of various types (such as images, videos, and points) with quite different properties. That can be an important reason for the lack of a survey on this topic. Therefore, \wang{as far as we are aware, this paper provides} the first survey on efficient visual recognition with DNNs. It targets at a systematic overview of recent advances and trends from various aspects, based on our expertise and experiences on major types of visual data, their various recognition models, and network compression algorithms.

\subsection{Related Surveys}
\label{subsec:related_surveys}

There are some related surveys published recently, but their scopes and contents are significantly different from ours. 

\ywunew{\emph{1) General introduction of DNNs and efficiency strategies for both model compression and hardware design}.
Sze et al. (2017) \cite{Sze_2017_PIEEE} provides one of the earliest surveys on general DNN models and their compression and hardware acceleration. Cheng et al. (2018) \cite{cheng_2018_FITEE} covers more advances on both network compression models (mainly CNNs) and hardware acceleration, but pays significantly more attention to the former. Though its scope is rather broad, the contents may be not detailed enough for many readers. Differently, Zhang et al. (2019) \cite{zhang_2019_Neurocomputing} has a much narrower coverage which can be good for the detailed directions if focuses on, but its distinctive categorization may confuse certain audience. Deng et al. (2020) \cite{Deng_2020_Survey}, one of the latest surveys, is comprehensive and professional in both DNN compression and hardware design. Conversely, understanding this survey completely is naturally a tough task.}

\ywunew{\emph{2)  General model compression and acceleration strategies (algorithm part)}.
This group include Cheng et al. (2018) \cite{cheng_2018_IEEESPM} which covers all major aspects of efficient DNNs but lacks the advanced contents, Lebedev et al. (2018) \cite{lebedev_2018_PAS} which is mainly about CNN-based models, and Elsken et al. (2019) \cite{Elsken_2019_NASSurvey} which focuses on the specific area of automatic network architecture search (NAS).}


\ywunew{\emph{3)  Hardware-based/Distributed  acceleration}.
 Guo et al. (2019) \cite{guo_2019_ATRT19} covers all aspects of FPGA-based DNN acceleration, especially on hardware-oriented model compression and efficient architectures for hardware design. The discussion is mainly on quantization and pruning, leaving tensorial methods and NAS out of its scope. Wang et al. (2019) \cite{wang_2019_ACMCompute} provides a more comprehensive survey of custom hardware oriented accelerators, including approximation methods for both convolutional and recurrent neural networks, but some promising approaches (\emph{e.g.}, neural architecture search) are still uncovered.
}

\ywunew{\emph{4)   Task-specific DNN models}.
Recently, a few surveys are focusing on the progresses on specific tasks, such as 3D data representation (Ahmed et al., 2019) \cite{ahmed_2019_arXiv}, texture representation (Liu et al., 2019) \cite{Liu_2019_Texture_Survey_IJCV}, generic object detection (Liu et al., 2019) \cite{Liu_2019_Detection_Survey_IJCV}. Though they have conducted comprehensive reviews on the existing models for such specific tasks, which is very valuable for understanding the progresses on the model development side, the efficiency issue is unfortunately not their focus and thus lacks sufficient coverage and in-depth analysis.
}

\subsection{Contributions and Organization}

\begin{figure*}
\centering
    \includegraphics[width=1.0\textwidth]{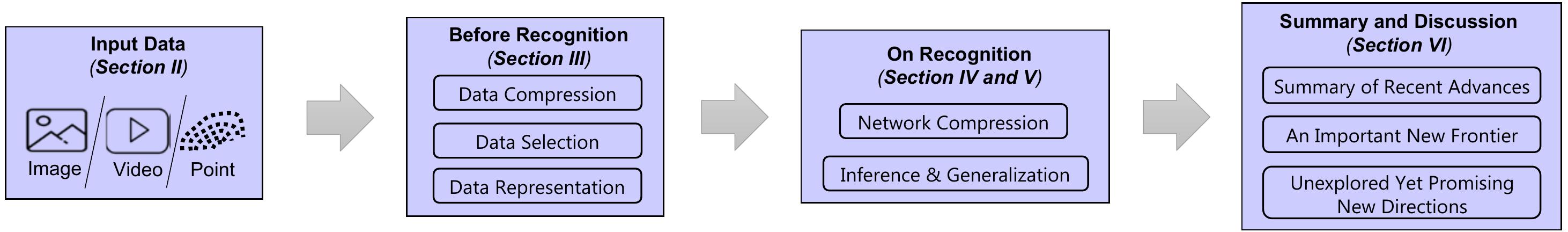}      
    \caption{The vein of this survey.}
\label{fig:vein}
\end{figure*}

By contrast, this survey mainly focuses on the global efficiency of the production line from the raw visual data to the final recognition results, and \wang{it is expected to} help the readers who are interested in the modern visual recognition tasks and their efficient DNN-based solutions. This paper contributes in the following aspects, which are also its novelties to the best of our knowledge.
\begin{enumerate}
\renewcommand{\labelenumi}{\theenumi)}
    \item A systematic survey of existing advances on efficient visual recognition approaches with DNNs, which is the first of its kind, as far as we are aware.
    \item The first summary on data related issues for efficient visual recognition, including data compression, data selection, and data representation.
    \item A new investigation of network compression models in the perspective of benefiting visual recognition tasks.
    \item A review of acceleration approaches for run-time inference and model generalization in the scope of efficient visual recognition.
    \item Insightful discussions on challenges, opportunities and new directions in efficient visual recognition with DNNs.
\end{enumerate}

For clearly making the vein of this survey, \wang{Fig. \ref{fig:vein} is drawn here} as the blueprint of the organization. Specifically, in Section \ref{sec:data_types}, we introduce the three main data types commonly concerned in visual recognition problems and discuss about their properties and the challenges related to each of them. Section \ref{sec:data_efforts} reviews the efforts on three aspects before the actual recognition part: data compression, data selection, and data representation. Section \ref{sec:network_compression} briefly introduces and analyzes the widely studied directions for network compression, within the scope of visual recognition. Section \ref{sec:generalization_inference} provides a summary of the recent progresses on efficient model generalization and fast inference at the testing phase, which are very important for real deployment of DNN-based visual recognition systems. Finally, Section \ref{sec:summary_discussions} overviews all efforts to generate a clear overall mapping, and discusses about some important uncovered aspects and new research directions.

\begin{figure*}
\centering
    \includegraphics[width=1.0\textwidth]{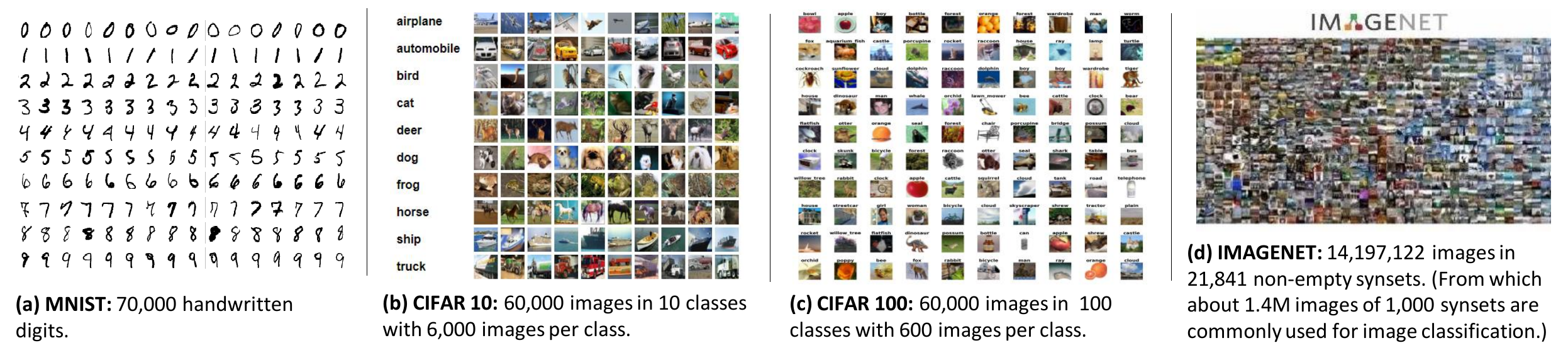}        
    \caption{The scale of images for recognition has been increased greatly.}
\label{fig:image_scale}
\end{figure*}

\section{Visual Recognition: Data Types, Challenges \wang{and Preliminaries}}
\label{sec:data_types}

Visual recognition covers several data types and a large number of detailed tasks\wang{, meanwhile, the recognition methods also include variant specific approaches. This section first introduces three most commonly concerned visual data types, i.e., image, video, and point, with their properties and recognition challenges. Then the related efforts before and on the recognition discussed in the whole paper are also simply listed as brief preliminaries for readers.}

\subsection{Data Types}

Images are the most widely studied visual data, probably due to their wide existence and relative simplicity of acquisition, storage, transmission, and processing. In many cases they are both efficient and sufficient for sharing information visually.
\wang{By contrast, as both the capturing devices and other related infrastructures and devices get greatly advanced and popularized, videos seem to be the most informative media. They have increased dramatically and will probably be even able to replace images in most scenarios soon. Videos appear very natural to humans, so they generally contain huge redundant information in the spatio-temporal domain. Points are relatively special since the radar and depth cameras appeared late, and they are usually sparse compared with images, but one dimension of them is higher and may have very large ranges. There are mainly two types of visual point data: point clouds and 2D/3D skeletons. The most valuable advantage of points is that they contain geometry information so the shape is the main information for recognition tasks.}

\subsection{Challenges}

\wang{It is clear that different data types have different characteristics, e.g., images: classical and efficient; videos: abundant informative and spatio-temporal information; points: sparse, high dimension and shape information. Thus the challenges they must deal with in recognition tasks should also be discussed.}

\subsubsection{Images: Larger scale and deeper understanding}\quad

\begin{figure*}
\centering
    \includegraphics[width=1.0\textwidth]{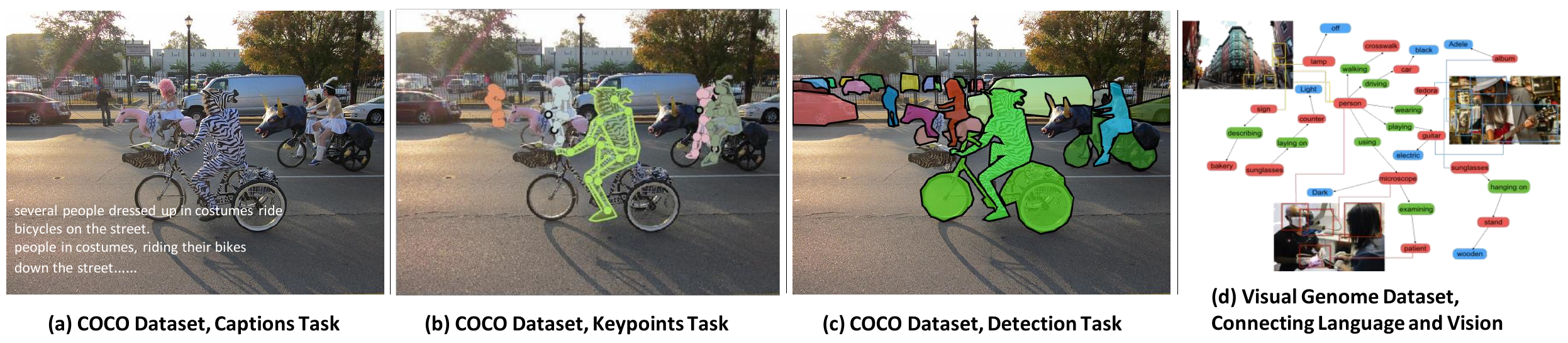}      
    \caption{The expected image recognition results are becoming deeper and richer, going far beyond simple classification.}
\label{fig:image_deeper_understanding}
\end{figure*}

There are two clear trends on image-based recognition with DNNs. \emph{The first is that the scale of processed data increases quickly}. As shown in Fig. \ref{fig:image_scale}, there is a clear history and trend that the benchmark dataset for developing new models have been shifting to larger scales and wilder contents (from MNIST \cite{Lecun_1998_MNIST_PIEEE} to CIFAR-10/CIFAR-100 \cite{Krizhevsky_2009_CIFAR10CIFAR100_MSThesis}, and to ImageNet \cite{imagenet_cvpr09}). With larger and more diverse training data, the trained DNN models can do more challenging recognition tasks (sometimes with the help of transfer learning), but it also brings greater challenges in efficient computation. \emph{The second is that the recognition is going toward deeper understanding and richer results}. Traditionally, classification or categorization is most commonly concerned, but recently a lot of efforts and progresses have gone far beyond that, spreading to many tasks including detection, attribute extraction, keypoint/pose estimation, semantic segmentation, image captioning, visual question answering, and even to the visual genome extraction, as shown in Fig. \ref{fig:image_deeper_understanding}. Such a trend greatly extends the research area and has attracted wider and stronger interests from both academia and industry. While new performance records are made on different tasks in much shorter time, the demands on exploring proper acceleration approaches become greater than ever, especially from the industry side which is eager to apply the latest models to various real scenarios.

\subsubsection{Videos: Information Richness and Sparsity}\quad

\begin{figure}
\centering
    \includegraphics[width=1.0\columnwidth]{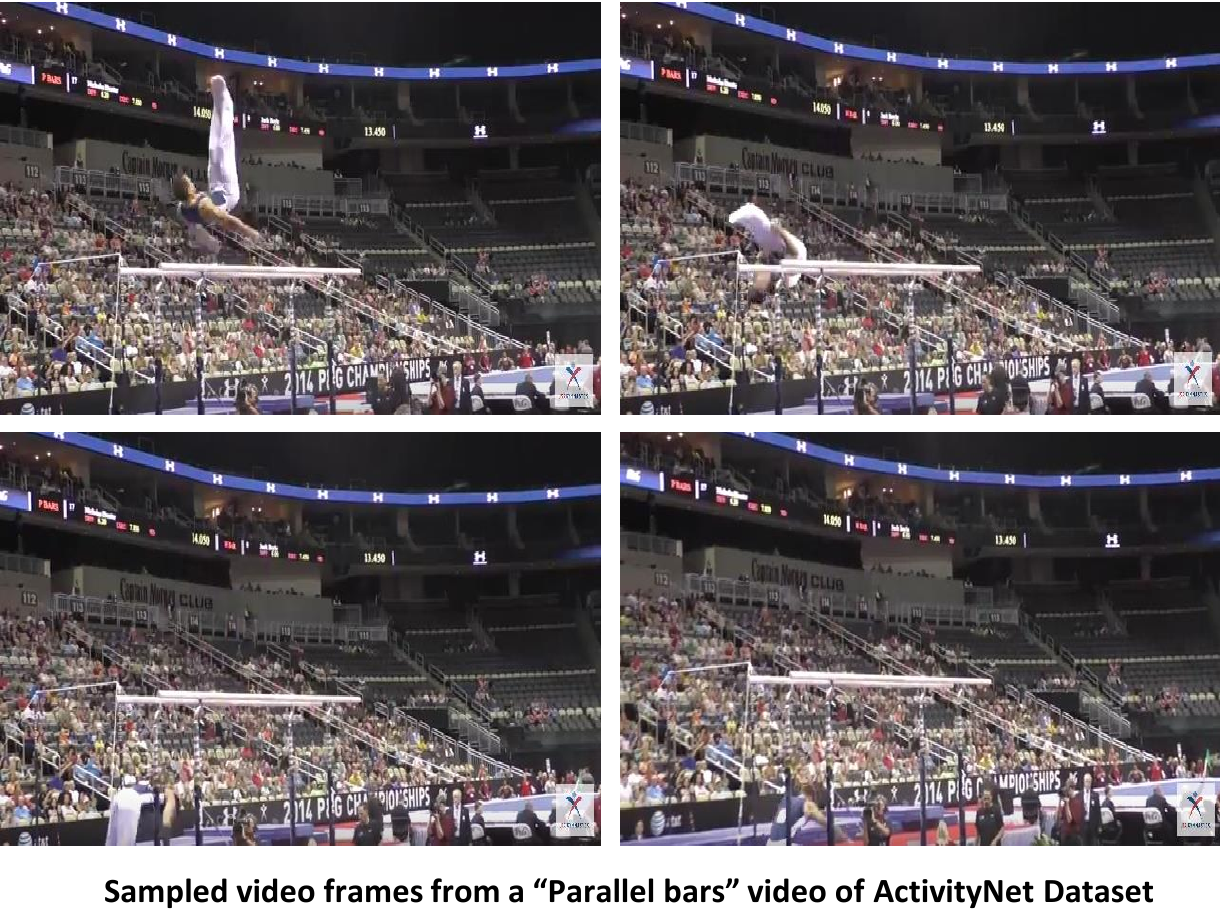}      
    \caption{Video recognition tasks usually target at extracting few high-level semantics (\emph{e.g.} category) from a large number of video frames, which are likely to have much redundant/irrelevant information.}
\label{fig:video_sparsity}
\end{figure}

\wang{Information richness is the natural challenge of videos compared with images. In contrast,} some of the visual challenges for images, such as occlusions and static background clutters, may get alleviated in videos when the motion information in the videos is properly treated. However, videos have their own \wang{particular} challenges. A major one is the sparsity of relevant information, as shown in Fig. \ref{fig:video_sparsity}. How to extract such sparse information from large amount of data without getting confused is a great challenge. Meanwhile, the bigger data size (compared with images) and in many cases the needs for real-time or even faster processing have made the efficiency issue even more important. Note that, most of the new recognition tasks for images also exist for videos, which require extra efforts for specific strategies on acceleration.

\subsubsection{Points: Geometry and Efficiency}\quad

\begin{figure}
\centering
    \includegraphics[width=1.0\columnwidth]{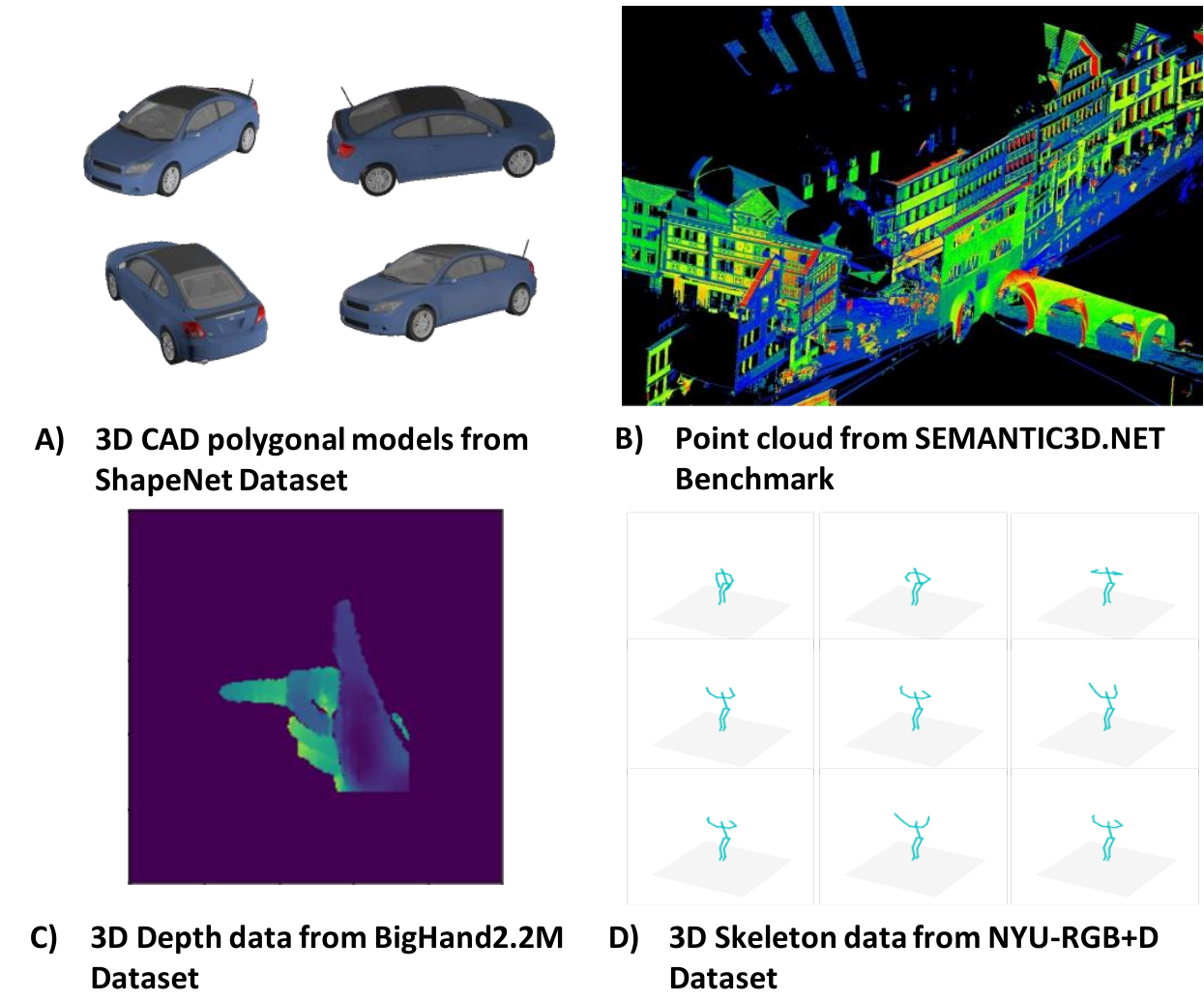}
    \caption{Examples of point data, in its general sense.}
\label{fig:points_examples}
\end{figure}

Though not as popular as images and videos, point data also play an important role in visual recognition. \wang{Except for the point clouds obtained from radar or depth camera (depth images can be easily turned into point clouds) for 3D object recognition \cite{xiang2016objectnet3d,zamir2018taskonomy}, and 2D/3D skeletons used for action recognition \cite{Jhuang_ICCV2013,Shahroudy_2016_CVPR,liu2017pku,tang2018deep,hou2018spatial}, 3D CAD models have also been used for visual recognition, such as the ShapeNet dataset \cite{Chang_2015_ShapeNet_arXiv}.} In a general sense, they can also be regarded as point data, as the CAD polygonal models can be tuned into voxels and point clouds when needed. Therefore, they are also included in the data samples as shown in Fig. \ref{fig:points_examples}. \wang{Since the geometry information in points is usually very informative for recognition, how to maintain such 3D geometry information whilst improving the efficiency of models is critical and is not easy as the dimensional range might be very large.}

\subsection{\wang{Preliminaries}}

\wang{In view of that varieties of methods will be introduced and discussed in the following content of this survey, it is necessary to make brief preliminaries to show most of the representative methods with their characteristics. Hence, Table \ref{Table-Preliminaries} is designed to afford a clear glance from data processing (Section \ref{sec:data_efforts}) to real deployment (Section \ref{sec:generalization_inference}), and the location of each method is also given. Please note that some trivial practices can not be exhibited here due to space limitation.}

\newcommand{\tabincell}[2]{\begin{tabular}{@{}#1@{}}#2\end{tabular}}
\begin{table*}
\caption{\wang{Preliminaries of the efficient methods mentioned in this survey.}}
\label{Table-Preliminaries}
  \centering
  \renewcommand\arraystretch{1.6}
  \resizebox{0.99\textwidth}{!}{
  \begin{tabular}{@{}@{\extracolsep{\fill}}c|c|c|c|c|c@{}}
  \hline
  \multicolumn{3}{c|}{Task} & Method & Characteristic & Location \\
  \hline
  \multirow{15}*{\tabincell{c}{Efforts\\on Data}} & \multirow{6}*{\tabincell{c}{Data\\Compression}} & \multirow{2}*{Images} & \tabincell{p{6cm}}{SIFT \cite{wu2013sift}, DCT domain \cite{gueguen2018faster}, Wavelet transform \cite{paul2015iris}} & \tabincell{p{6cm}}{Greatly optimized and easy to be used directly} & Section \ref{sec:data_efforts}-A1 \\
  \cline{4-6} & & & \tabincell{p{6cm}}{Encoder-decoder networks \cite{Rippel2017Real,Ball2018Variational,Minnen2018Image,NIPS2018_8275}} & \tabincell{p{6cm}}{Ensure both the relevant information and speed} & Section \ref{sec:data_efforts}-A1 \\
  \cline{3-6} & & \multirow{3}*{Videos} & \tabincell{p{6cm}}{H.265 \cite{sullivan2012overview} and H.264 \cite{wiegand2003overview}} & \tabincell{p{6cm}}{Can not be optimized end-to-end} & Section \ref{sec:data_efforts}-A2 \\
  \cline{4-6} & & & \tabincell{p{6cm}}{Coding by CNN \cite{chen2017deepcoder}, Kalman filtering network \cite{Lu_2018_ECCV}, frames reconstructed by CNN+LSTM \cite{wu2018video}} & \tabincell{p{6cm}}{Must restore the compressed data to a raw video format and bring lots of extra cost} & Section \ref{sec:data_efforts}-A2 \\
  \cline{4-6} & & & \tabincell{p{6cm}}{Deep feature flow \cite{Zhu_2017_CVPR}, direct training \cite{wu2018compressed}} & \tabincell{p{6cm}}{End-to-end and real-time process} & Section \ref{sec:data_efforts}-A2 \\
  \cline{3-6} & & Points & \tabincell{p{6cm}}{Auto-encoder-based geometry codec\cite{Yan2019Deep,wang2019learned,Yang2017FoldingNet}} & \tabincell{p{6cm}}{Aiming at geometric characteristics} & Section \ref{sec:data_efforts}-A3 \\
  \cline{2-6} & \multirow{4}*{\tabincell{c}{Data\\Selection}} & Images & \tabincell{p{6cm}}{Subsampling by active learning \cite{ImageSel_ADS_arXiv19,ImageSel_Core_Set_ICLR18}, gradient method \cite{ImageSel_EmpStudy_arXiv18}, or clustering \cite{ImageSel_Redundancies_10P_arXiv19}} & \tabincell{p{6cm}}{Noises and redundant samples can be reduced to prevent overfitting} & Section \ref{sec:data_efforts}-B1\\
  \cline{3-6} & & \multirow{2}*{Videos} & \tabincell{p{6cm}}{Random key frames selection \cite{Gao_2017_ICCV,carreira2017quo,Xie_2018_ECCV,zolfaghari2018eco}} & \tabincell{p{6cm}}{Minimum computation cost but missing information} & Section \ref{sec:data_efforts}-B2 \\
  \cline{4-6} & & & \tabincell{p{6cm}}{Key frames prediction by reinforcement learning \cite{yeung2016end,huang2018sap,lan2018ffnet,fan_watching_2018_IJCAI}, adaptive pooling \cite{kar_2017AdaScan_CVPR}, MA LSTM \cite{Wu_2019_CVPR}} & \tabincell{p{6cm}}{Select high-quality key frames and avoid to process redundant frames} & Section \ref{sec:data_efforts}-B2 \\
  \cline{3-6} & & Points & \tabincell{p{6cm}}{Sampling with modeling attention \cite{PointSel_Yang_2019_CVPR,PointSel_Paigwar_2019_CVPR_Workshops,PointSel_KingkanOH_BMVC18}} & \tabincell{p{6cm}}{Superiority over traditional or neural methods} & Section \ref{sec:data_efforts}-B3 \\
  \cline{2-6} & \multirow{5}*{\tabincell{c}{Data\\Representation}} & Images & \tabincell{p{6cm}}{Pre-trained model on other datasets \cite{ImageRep_PreTrained_TMI16,ImageRep_PreTrained_CBM17,ImageRep_Unsupervised_arXiv19,ImageRep_CNN_Features_Baseline_CVPRW14}} & \tabincell{p{6cm}}{Make the entire training process efficient} & Section \ref{sec:data_efforts}-C1 \\
  \cline{3-6} & & \multirow{2}*{\tabincell{c}{Videos}} & \tabincell{p{6cm}}{I3D \cite{carreira2017quo}, 2DCNN + 1D temporal convolution \cite{Wu_2018T_CN_AAAI}} & \tabincell{p{6cm}}{Frames representation and temporal correlations} & Section \ref{sec:data_efforts}-C2 \\
  \cline{4-6} & & & \tabincell{p{6cm}}{Dynamic image \cite{bilen_2016DynamicImage_CVPR,bilen_2018DynamicImage_TPAMI}} & \tabincell{p{6cm}}{Turn a whole video into one single informative image} & Section \ref{sec:data_efforts}-C2 \\
  \cline{3-6} & & Points & \tabincell{p{6cm}}{Align the coordinates dimension \cite{yang2019ddnet,li2018co}} & \tabincell{p{6cm}}{One dimension is reduced} & Section \ref{sec:data_efforts}-C3 \\
  \hline
  \multirow{10}*{\tabincell{c}{Network\\Compression}} & \multirow{3}*{\tabincell{c}{Compact\\Networks}} & CNNs & \tabincell{p{6cm}}{Light receptive field \cite{szegedy2016rethinking,wang2018understanding}, topology \cite{Iandola2017squeeze,zhang2018shufflenet}, or block \cite{CompactNet_PNN_juefei-xu2018pnn,CompactNet_LBCNN_Juefei-Xu_2017_CVPR}} & \tabincell{p{6cm}}{Some compact designs become standard neural structure such as bottleneck and depthwise convolution} & Section \ref{sec:network_compression}-A1 \\
  \cline{3-6} & & RNNs & \tabincell{p{6cm}}{Simpler units \cite{wu2016investigating,van2018unreasonable} or architectures \cite{sak2014long,wu2016google}} & \tabincell{p{6cm}}{Hard to implement, limited compression ratio} & Section \ref{sec:network_compression}-A2 \\
  \cline{3-6} & & NAS & \tabincell{p{6cm}}{Reinforcement learning \cite{zoph2017neural}, evolutionary algorithm \cite{real2018regularized}, bayesian optimization \cite{Kandasamy_NASBO}, gradient-based \cite{cai2018proxy}} & \tabincell{p{6cm}}{Can surpass human designs in both accuracy and efficiency, promising but still needs further studies} & Section \ref{sec:network_compression}-A3 \\
  \cline{2-6} & \multirow{2}*{\tabincell{c}{Tensor\\Decomposition}} & Tucker & \tabincell{p{6cm}}{CP-CNN \cite{AstridCPCNN}, Tucker-CNN \cite{ChienTuckerLayer}, BTD-LSTM \cite{Ye_2018_BTD}} & \tabincell{p{6cm}}{Curse of dimensionality and complex computation} & Section \ref{sec:network_compression}-B1 \\
  \cline{3-6} & & \tabincell{c}{Tensor\\Network} & \tabincell{p{6cm}}{TT-CNN \cite{NovikovTT,GaripovTTCNN,Wang_2020_TT3DCNN}, TT-RNN \cite{TjandraTTRNN1,YangTTRNN}, TC-RNN \cite{PanTCRNN}, HT-RNN \cite{Wu_2020_Hybrid,Yin_2020_HTRNN}} & \tabincell{p{6cm}}{High compression ratio, \emph{in situ} training, hard to avoid accuracy loss} & Section \ref{sec:network_compression}-B2 \\
  \cline{2-6} & \multirow{2}*{\tabincell{c}{Data\\Quantization}} & Projection & \tabincell{p{6cm}}{WAGE \cite{Wu_2018_WAGE}, full 8-bit training \cite{Yang_2019_WAGEU}} & \tabincell{p{6cm}}{Project floats to distributed integers, mainstream way} & Section \ref{sec:network_compression}-C1 \\
  \cline{3-6} & & Optimization & \tabincell{p{6cm}}{XNOR-NET \cite{rastegari2016xnor}, AutoQ \cite{Lou_2020_AutoQ}} & \tabincell{p{6cm}}{More attention to the whole network} & Section \ref{sec:network_compression}-C1 \\
  \cline{2-6} & \multirow{2}*{\tabincell{c}{Pruning}} & Search & \tabincell{p{6cm}}{Low-precision estimation \cite{lin2017predictivenet,song2018prediction}, negative activation prediction \cite{aklaghi2018snapea}} & \tabincell{p{6cm}}{Vast computing time, extra indices of pruned weights or neurons} & Section \ref{sec:network_compression}-D1 \\
  \cline{3-6} & & Optimization & \tabincell{p{6cm}}{Structured sparsity \cite{wen2016learning}, ThiNet \cite{luo2017thinet}, SSR \cite{Lin_2020_SSR}} & \tabincell{p{6cm}}{Adaptive to large DNNs, structured pruning} & Section \ref{sec:network_compression}-D1 \\
  \cline{2-6} & \multicolumn{2}{c|}{Joint Compression} & \tabincell{p{6cm}}{Decompose + quantize \cite{liu2015sparse,wen2017coordinating}, quantize + prune \cite{han2015deep,choi2018compression}} & \tabincell{p{6cm}}{Extremely high compression ratio, maintaining accuracy is critical} & Section \ref{sec:network_compression}-E2 \\
  \hline
  \multirow{4}*{\tabincell{c}{Inference \&\\Generalization}} & \multirow{2}*{\tabincell{c}{Fast\\Inference}} & Data-aware & \tabincell{p{6cm}}{Recurrent residual module \cite{pan_2018_CVPR}, efficient inference engine \cite{han_2016eie_ISCA}, scale-time lattice \cite{Chen_2018_CVPR}} & \tabincell{p{6cm}}{Efforts on reducing the computation on the redundant data are important for inference} & Section \ref{sec:generalization_inference}-A1 \\
  \cline{3-6} & & Network-centric & \tabincell{p{6cm}}{Prune whole blocks \cite{chang_2019urnet_arXiv}, dynamic compression ratio \cite{fang_2018nestdnn_MobiCom}, integrate resource and input \cite{mullapudi_2018_CVPR}} & \tabincell{p{6cm}}{General network-centric compression for fast inference should be evaluated by the proposed KPI} & Section \ref{sec:generalization_inference}-A2 \\
  \cline{2-6} & \multirow{2}*{\tabincell{c}{Generalization via\\transfer learning}} & Fine-tuning & \tabincell{p{6cm}}{Transfer the learned knowledge of the source task to a related task with domain adaptation \cite{qiao2018deep,rad2018feature,watanabe2018multichannel,saito2018strong,chen2017show}} & \tabincell{p{6cm}}{Pre-trained models with fine-tuning is generally better than training-from-scratch for most target tasks with smaller-scale datasets} & Section \ref{sec:generalization_inference}-B \\
  \cline{3-6} & & Fast transfer & \tabincell{p{6cm}}{Adapt the pre-trained model to the target task together with compression \cite{Transfer_Rosenfeld_2018_PAMI,Transfer_Rebuffi_2018_CVPR,Transfer_houlsby19a_2019_ICML}} & \tabincell{p{6cm}}{Learning fewer parameters during transferring is more efficient than naive full model fine-tuning} & Section \ref{sec:generalization_inference}-B \\
  \hline
  \end{tabular}}
\end{table*}


\section{Before Recognition: Efforts on the Data}
\label{sec:data_efforts}

Visual data have their own properties, which can be made use of when efficient recognition models are designed. Much efforts can be made to change/map the data to a more compact form before applying the recognition models. These efforts may be grouped into three categories based on their functionalities: \textbf{data compression} \emph{(reducing redundancy and irrelevance)}, \textbf{data selection} \emph{(reducing irrelevance)}, and \textbf{data representation} \emph{(increasing compactness)}. In this section, we overview and summarize existing progresses in all these three aspects, and organize the contents for each of them according to their related data types. By doing so, their motivations and strategies can be easily understood and searched for.

\subsection{Data Compression}

\begin{figure}
\renewcommand{\arraystretch}{0.4}
\centering
    \includegraphics[width=0.5\textwidth, height=0.15\textwidth]{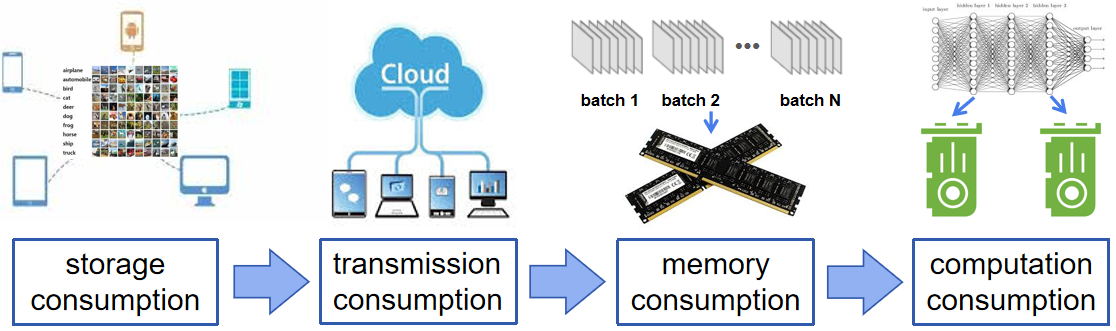}        
    \caption{Major types of consumption in deep learning}
\label{fig:consumption}
\end{figure}

As datasets and networks grow in size, the great memory consumption and high computational complexity have made the training of deep neural networks a challenge and hindered the popularity of AI. Specifically, for deep learning, there are four types of consumption (as shown in Fig. \ref{fig:consumption}) due to large data sets: a) \emph{\textbf{storage} consumption}, most of the commonly used datasets are now hundreds of gigabytes in size, the situation of which puts a lot of pressure on storage hardware; b) \emph{\textbf{transmission} consumption}, the transmission of large amounts of image/video data can be a challenge to network bandwidth; c) \emph{\textbf{memory} consumption}, when training a network, usually the larger the batch size (the amount of data fed into the network each time) is, the better the performance is, and thus a larger memory is always desired; d) \emph{\textbf{computing} consumption}, DNNs usually need to rely on powerful computing resources (\emph{e.g.} a GPU), especially when large datasets have to be handled, and such a demand has a rising trend. Increasing data size may lead to a better recognition performance, but it can also increase all these four types of consumption.

To reduce the consumption needs, the commonly used and also very important and effective strategy is to compress the data, which can be done before the recognition task and it is relatively less task-dependent.

\subsubsection{Image compression}\quad

\begin{figure}
\centering
    \includegraphics[width=1.0\linewidth]{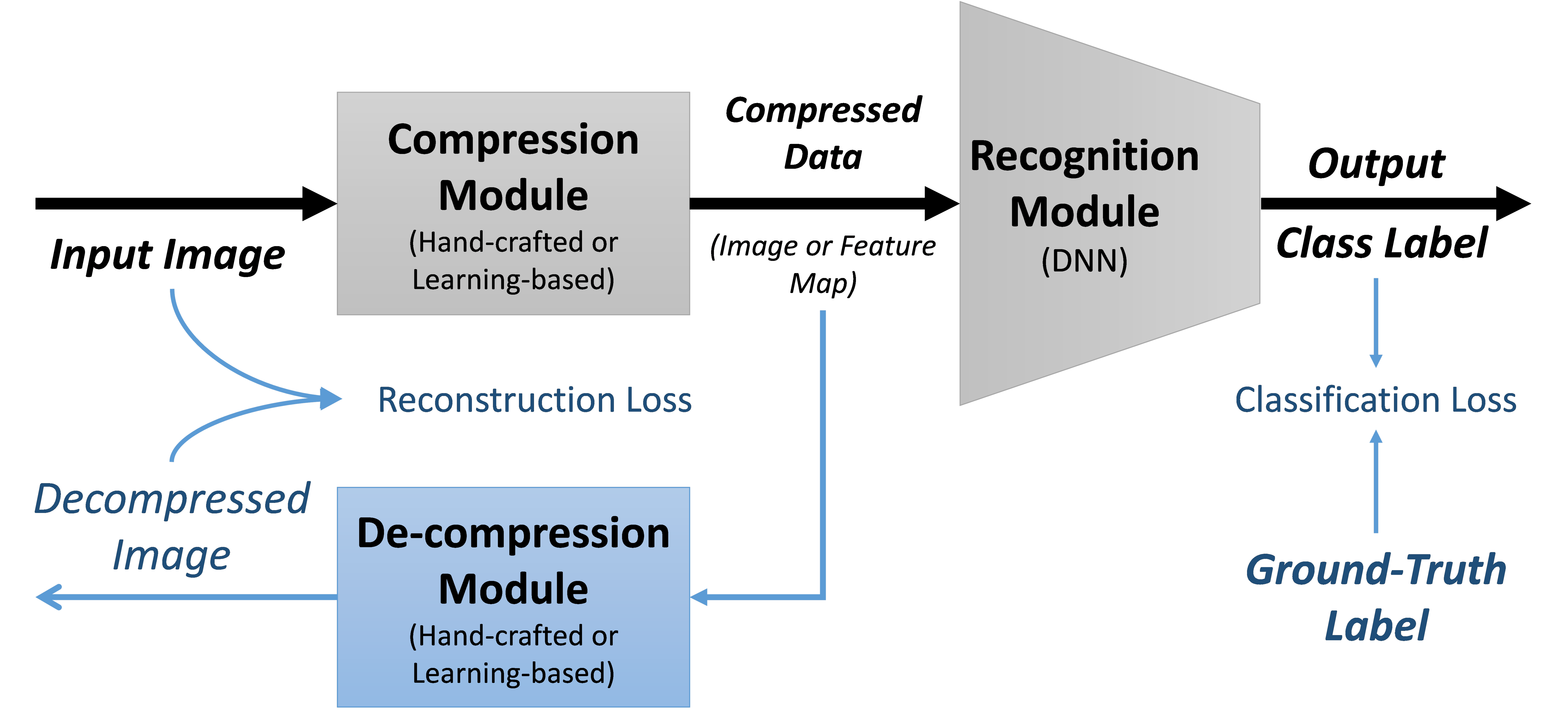}        
    \caption{A general framework for joint image compression and recognition. The upper row in black shows the inference (test) phase, while the whole figure (including the blue parts) describes the training phase. Note that the compression module has to be optimized for minimizing both the reconstruction loss and the classification loss, different from the recognition module which is only optimized for the later one.}
\label{fig:compression_recognition_framework}
\end{figure}


Image compression and image recognition can be linked together to save the consumption on image decompression, which has been proved to be more efficient and in many cases can be made to be more effective as well. A general framework for decompression-free joint compression and recognition is shown in Fig. \ref{fig:compression_recognition_framework}, which illustrates both training phase and test/inference phase. While the recognition module in the framework is generally DNN-based in the scope of this survey, the compression module can be either hand-crafted or learning-based. Though \wang{many researchers} may expect a purely learning-based (DNN-based) framework, the current reality is that the sophisticated hand-crafted image compression models are still more popular. It is not easy to invent a new DNN-based compression model that outperforms the greatly optimized hand-crafted models. For researchers who know much more about recognition than compression, usually they tend to keep the compression module as it is and put more efforts on how to make their recognition modules better to receive the compressed data. Therefore, in this subsection, \wang{this survey} still starts from introducing the hand-crafted compression models before going to the learning-based models, but putting more efforts on the later to encourage more future researches on inventing better purely learning-based solutions.

\begin{figure*}
\renewcommand{\arraystretch}{0.4}
\centering
    \includegraphics[width=1.0\textwidth, height=0.22\textwidth]{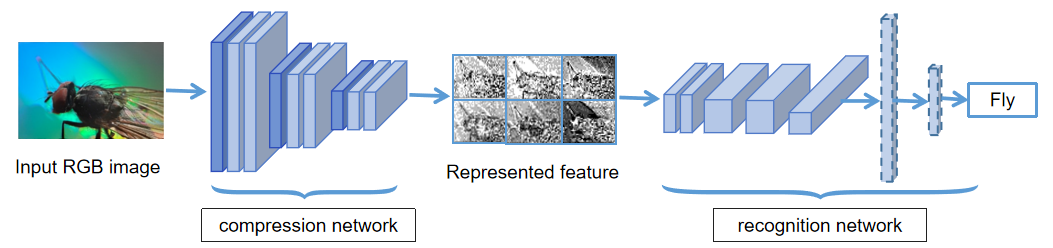}        
    \caption{The input image is represented as a series of compressed feature maps, and subsequent recognition network learns classification information from these feature maps. }
\label{fig:learning}
\end{figure*}

\noindent \textbf{Hand-crafted Lossy Compression}.
Although several more complicated compression standards have been developed including JPEG, WebP and BPG, JPEG remains the most widely used one for lossy image compression. Recently, a lot of interesting works \cite{wu2013sift,ehrlich2018deep,liu2018deepn,gueguen2018faster,javed2018review,paul2015iris} have been developed to learn the feature distribution directly from compressed data. Wu et al. \cite{wu2013sift} formulate the popular SIFT feature extraction in JPEG's DCT domain, which indicates that effective visual features can be directly extracted from the compressed data. Ehrlich et al. \cite{ehrlich2018deep} introduce a general method to learn a Residual Network in the JPEG transform domain. Liu et al. \cite{liu2018deepn} combine compression and recognition based on JPEG's transformation, and it has been proved by experiments that it could achieve 3.5$\times$ compression rate improvement, while its consumption is only 30\% of the conventional JPEG without classification accuracy degradation. In \cite{gueguen2018faster}, it is proposed to learn DNN parameters directly from the DCT coefficient of JPEG compression,  which is $ 1.77\times$ faster than ResNet-50 at the same accuracy. Javed et al. \cite{javed2018review} propose a method for reviewing document images, which is directly based on the compressed documents data. In \cite{paul2015iris}, Haar wavelet transform is utilized to compress and decompress high-resolution iris images, enabling fast and accurate iris recognition.\\      
 
\noindent \textbf{Learning Based Lossy Compression}.
Although traditional compression algorithms are carefully constructed, there is still room for the improvement in compression efficiency. For example, the conversions are fixed and cannot be adaptive to fit different inputs. In addition, pre-defined quantization implementations can result in data redundancy. Moreover, limited by manual design, the algorithms are usually hard to be optimized for a specific metric, even if the metric is a perfect assessment of image reconstruction quality. Therefore, more attentions have been paid to learning based compression methods in recent years. Different from traditional methods, in a learning-based approach, the parameters of the neural network are automatically learned from a large amount of data by definite optimization objectives to deal with specific situations. A general pipeline is that the input image $x$ is firstly processed by the analysis network (encoder) to generate compressed feature-maps, which are then converted into a set of bit-streams by quantization and lossless arithmetic coding. After that, they are used to generate a recovered image $\hat{x}$ by a re-factoring network (decoder), and the entire network is trained end-to-end until convergence. On the basis of this pipeline, many excellent image compression methods \cite{Rippel2017Real,Ball2018Variational,Minnen2018Image,NIPS2018_8275} have been proposed. 
 
\noindent \textbf{Joint Compression and Recognition}.
Stimulated by the tremendous memory reduction caused by compression algorithms, some methods \cite{oyallon2018compressing,torfason2018towards,Chang2019RandNetDL,1909.05638} combine compression with recognition to improve effectiveness and efficiency. Specifically, in order to save the decompression consumption, a well learned analysis network can be directly cascaded with a downstream recognition network as shown in Fig. \ref{fig:learning}. 
Such a joint optimization not only tries to retain as much classification relevant information as possible during compression, but also accelerates the speed of inference and optimizes the consumption of training compared with models directly trained on the input images. Detailed models can be found in \cite{oyallon2018compressing,torfason2018towards,Chang2019RandNetDL,1909.05638}. 

\subsubsection{Video compression}\quad

As the most common media, video was said to take more than $70\%$ of all Internet traffics according to the white paper of ``Cisco Visual Networking Index: Forecast and Methodology, 2016-2021'', and now the percentage is probably even greater. In the past few years, many representative progresses \cite{ding2018trunk,pigou2018beyond,ullah2018action,Tulyakov_2018_CVPR,sun2018optical} have been made in video-based recognition. These methods, however, all focus on designing a special neural network for analyzing frames, ignoring the fact that videos are in a compressed format during transmission and storage. Therefore, extra time and storage are needed for decompression before the analysis. In order to improve the effectiveness and efficiency of video recognition, it is necessary to apply efficient compression and decompression beforehand or directly use compressed data for recognition.\\


\noindent \textbf{Compression Algorithms}.
As the most popular video compression algorithms, some highly efficient compression standards such as HEVC(H.265) \cite{sullivan2012overview} and AVC(H.264) \cite{wiegand2003overview} have been in use for a long time. Taking the H.264 algorithm as an example, three kinds of frames are defined in the encoding protocol. The fully encoded frame is called I-frame (key frame), and the frame containing only the difference partial encoding generated by the I-frame is called P-frame. The highly compressed frame obtained by using both previous and forward frames for data reference is called B-frame. There are two core algorithms used by H.264: intra-frame compression and inter-frame compression. Among them, intra-frame compression is an algorithm for generating I-frames, and inter-frame compression can generate highly compressed B-frames and P-frames.

\indent To achieve inter-frame compression, H.264 relies on many handcrafted modules, such as DCT transform module, block-based motion estimation module, and motion compensation module. Although these modules are well designed, they are not optimized end-to-end.

\indent Recently, several DNN-based video compression methods have been proposed for intra prediction \& residual coding \cite{chen2017deepcoder}, post-processing for predicted frames \cite{Lu_2018_ECCV}, inter-frame interpolation \cite{wu2018video},  and full network-based video compression \cite{lu2018dvc,1908.05717}. Chen et al. \cite{chen2017deepcoder} design two convolutional neural networks to encode predicted images and residual images, respectively, and the reliable experimental results prove that deep neural networks can achieve better results than hand-crafted modules. Lu et al. \cite{Lu_2018_ECCV} model the video artifact reduction task as a Kalman filtering procedure and restore decoded frames through a deep Kalman filtering network. By constructing a recursive filtering scheme based on the Kalman model, more accurate time information can be used to obtain better reconstruction quality. Wu et al. \cite{wu2018video} regard the video compression challenge as a repeated image interpolation challenge, so that the remaining frames can be reconstructed from the key frames through an interpolation reconstruction network. In addition to this, their algorithm provides a compressible code to disambiguate different interpolations and encode key frames as faithfully as possible. However, although the interpolation network is end-to-end optimized, motion information still requires additional calculations, which depend in part on other algorithms. In \cite{lu2018dvc}, an end-to-end video compression deep model that jointly optimizes all the components for video compression has been proposed. Specifically, the optical flow estimation network is used to obtain motion information, and the compressed network is used to compress both motion information and residuals. These two different networks are jointly learned through end-to-end optimization. Habibian et al. \cite{1908.05717} present a depth generation model for lossy video compression, which consists of a three-dimensional automatic encoder with discrete potential space and an autoregressive prior for entropy coding. Self-encoder and transcendental encoder are trained jointly to achieve the best rate-distortion curve. This method is superior to the latest learning video compression network based on motion compensation or interpolation. In addition, three extension directions are proposed: semantic compression, adaptive compression and multimodal compression. These directions mentioned above will undoubtedly lead to new video compression applications, which may be realized by DNNs but not classical codecs.

\noindent \textbf{Utilizing Compressed Data}.
Due to the time redundancy and the enormous size of data streams, there is a large amount of redundant or irrelevant information in the video data, which makes learning neural networks difficult and slow. Therefore, most video recognition algorithms use a video compression algorithm such as H.264 as pre-processing, which can reduce superfluous information by two orders of magnitude. In recognition, the mainstream method is to restore the compressed data to a raw video format and then handle each frame as a RGB image. Considering that neural networks can learn features from data, the process of decoding compressed data may be skipped. 

\begin{figure}
\renewcommand{\arraystretch}{0.4}
\centering
    \includegraphics[width=0.5\textwidth, height=0.4\textwidth]{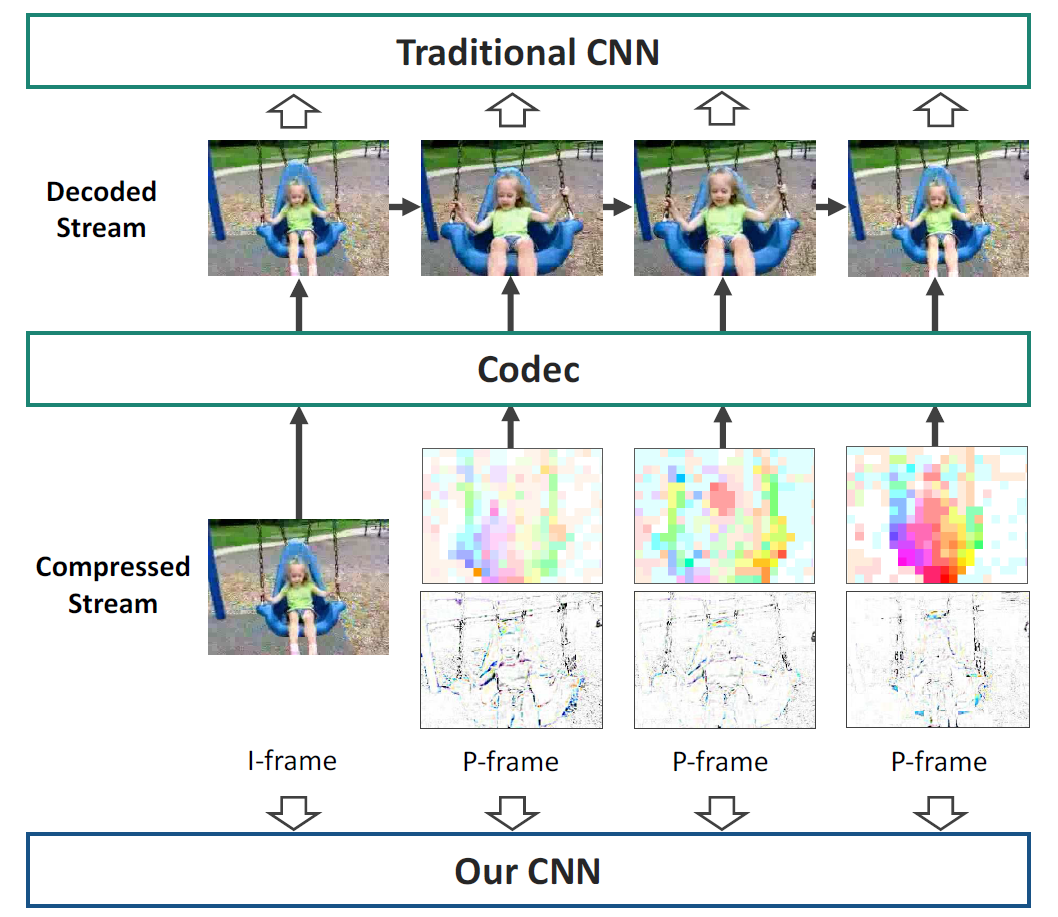}        
    \caption{A recently proposed video recognition network, which was directly trained on compressed videos \cite{wu2018compressed}.}
\label{fig:cvar}
\end{figure}

Zhu et al. \cite{Zhu_2017_CVPR} present a fast and accurate video recognition framework namely deep feature flow. It extracts depth features on key frames (I-frames) through a convolutional network and maps them to other frames for auxiliary prediction through the flow field. In this process, significant efficiency gains can be achieved by reducing the amount of computation. In \cite{wu2018compressed}, a network trained directly on compressed videos is proposed, as shown in Fig. \ref{fig:cvar}. The benefits of this design are three-fold. Firstly, the compressed video representation removes large amounts of redundant information and preserves useful motion vectors. Secondly, compressed video representations are more convenient for exploring video correlation than individual images. Finally, such an approach is more efficient because only informative signals are processed rather than near-duplicates, and the efficiency can also be improved by skipping the steps of decoding the video as the video is stored in a compressed version. 

In general, video compression technology has already played a significant role in video-based recognition tasks, no matter whether it is about a traditional video compression standard or a network-based learning algorithm. In most user-oriented application scenarios, such as public security monitoring and real-time scene replacement, real-time processing is usually a must and stability of the algorithm is a desire. In those cases, video compression has been proved to have great importance and  application potential.
 
\subsubsection{Point Cloud Compression}\quad

Different from image/video data, point cloud contains more complex structures, which make the compression much more difficult. The most efficient way of compressing the point cloud is to utilize its geometric characteristics. Quach et al. \cite{Quach2019Learning} present a novel data-driven geometry compression method for static point clouds based on learned convolutional transforms and uniform quantization. The first auto-encoder-based geometry compression codec is proposed in \cite{Yan2019Deep}, where the point cloud is treated as input rather than voxel grids or collections of images. Inspired by the great success of variational auto-encoders (VAE) in image/video compression, Wang et al. \cite{wang2019learned} present a DNN based end-to-end point cloud geometry compression framework, in which the point cloud geometry is first voxelized, scaled and partitioned into non-overlapped 3D cubes which is then fed into a 3D convolutional networks for generating the latent representation. Furthermore, Yang et al. \cite{Yang2017FoldingNet} present a novel and interesting end-to-end deep auto-encoder to achieve the state-of-the-art performance for supervised learning tasks on point clouds. In this work, a graph-based enhancement is firstly enforced (in the encoder) to promote local structures and form a low dimensional codeword, which can be used to deform/fold a canonical 2D grid and then reconstruct the 3D object surface of the input point cloud. The learned encoder can be treated as a compression module, which has been proved to have great generalization abilities: experiments on major datasets showing that the folding can achieve higher classification accuracy than other unsupervised methods with significantly fewer parameters (7\% parameters of a decoder with fully-connected neural networks). The research on learning point cloud compression has just started and got tested on recognition tasks, and this new topic remains largely unexplored. Encouraged by the successes of these pioneering works, more and bigger progresses can be expected in the coming near future.

\subsection{Data Selection}

Selecting only relevant and informative parts from the raw data for recognition can significantly reduce the computational cost and may also lead to a better recognition accuracy (as disturbance by irrelevant information/noise is reduced). Due to the great differences between the data types in terms of data structure and information characteristics, the researches focus and method on them are also quite different from each other. For image data, sample selection for training is the main concern. In the case of video data, frame selection inside each video (for both training and testing) matters most as continuous video frames contain a lot of redundant information. Similarly, point data also have much redundancy and subset sampling is the main stream. Details on the motivations, strategies and methods for them are given below.

\subsubsection{Image Data}\quad

Currently, DNNs are generally data hungry, namely, the more labeled training data the better performance they can achieve. However, more data also means more costs, which include the efforts for data acquisition and labeling and all the consumption (storage, transmission, memory, and computation) for learning. Given a fixed amount of training data, directly scaling up the computation (by increasing the number of parameters and/or doing more iterations) usually has a performance upper bound and more computation after that goes more towards \emph{overfitting}. Such an overfitting is believed to be the result of \emph{the inherent noise} (or certain \emph{redundant samples} in a softer tone) in the training data \cite{ImageSel_ADS_arXiv19}. Therefore, reducing such noise or redundant samples shall not only saves consumption for model learning, but also have the potential to even boost the model's performance.

Subsampling the training data has recently been studied under such a motivation, and it has already shown quite promising results in the past couple of years. Since so far such researches have only concerned about image recognition and subsampling can be regarded as data selection, we introduce them here. In a more general sense, these approaches shall also be applicable or at least have the potential to be made applicable for other data types. Since DNNs prefer more data, naive random subsampling likely ends up with inferior performance. Therefore, some efforts have focused on how to get subsets better than randomly sampled ones or revealing the inequality of training samples. Core-set selection \cite{ImageSel_Core_Set_ICLR18} and representative subset finding \cite{ImageSel_EmpStudy_arXiv18} are good examples in this direction. However, doing better than random subsampling does not guarantee no drop in performance. More recent works improve over these by identifying redundant samples for removing them without sacrificing the recognition performance. Clustering in the DNN feature space \cite{ImageSel_Redundancies_10P_arXiv19} shows a successful removal of 10\% semantically redundant samples from CIFAR-10 and ImageNet datasets, while a slightly later work on ``select via proxy'' is able to push this reduction to 40\% on CIFAR-10 with no performance loss with the help of three uncertainty metrics. The very recent work on Active Dataset Subsampling (ADS) \cite{ImageSel_ADS_arXiv19} presents even more encouraging results: removing 50\% of CIFAR-10 training samples yet even perform slightly better than training on the full dataset. Similarly, on ImageNet dataset they are able to outperform training with all data by using only 80\% of it.

As shown in \cite{ImageSel_ADS_arXiv19}, advantages of subsampling are not limited to its abilities to maintain or even improve recognition performance whilst saving all learning consumption. The sampled subset can be effective for other model architectures which are not trained on. Moreover, the subsampling method itself (\emph{e.g.} ADS) may be applicable to many tasks and various data size settings. The research in this direction has just begun and there is still large room for exploring. An important and highly valuable problem is how to do that with noisy or weak labels as collecting high-quality labels is usually a great challenge.

\subsubsection{Video Data}\quad

\begin{figure}
\renewcommand{\arraystretch}{0.4}
\centering
    \includegraphics[width=0.7\linewidth]{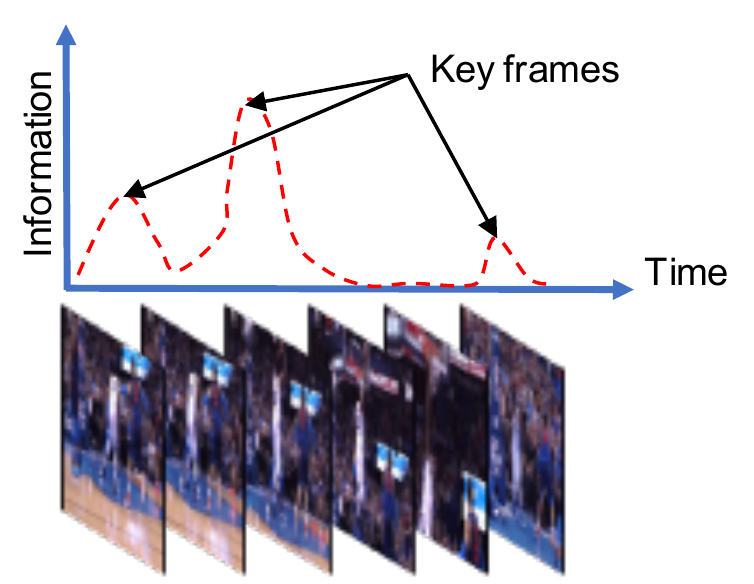} 
    \caption{Video key frames which are informative for recognition could be sparse and unequally located.}
\label{fig:key_frame}
\end{figure}

\wang{Since the huge redundant information in the spatio-temporal domain is prime for videos, instead} of performing expensive processing on every frame to approach target tasks, such as object detection and action recognition, selecting key frames and performing the major processing sparsely is a more efficient choice (see Fig. \ref{fig:key_frame} for the motivation). However, how to efficiently select proper key frames remains an open issue. Since relevant and discriminative video information could be unequally located in its temporal domain, obtaining few high-quality key frames without loosing those information could cost significant extra computation. How to do that efficiently is an important issue for video-based recognition tasks.

Generally, existing approaches on selecting key frames could be divided into two types. The first one is to randomly pickup key frames by predefined intervals. It is commonly applied in video recognition tasks, which may also include temporal boundary detection~\cite{Gao_2017_ICCV,carreira2017quo,Xie_2018_ECCV,zolfaghari2018eco}. Although this method takes the minimum computation cost, picking up frames by a large interval may miss critical information, while using a small interval could increase the post-hoc computation costs. Alternatively, in another type of approaches, the information from processed frames can be used to predict the possible key frames in the future, which could avoid to process many redundant frames and select high-quality key frames~\cite{yeung2016end,kar_2017AdaScan_CVPR,huang2018sap,lan2018ffnet,Wu_2019_CVPR}. The sampling and early stopping can be learned by an end-to-end deep reinforcement learning model as studied in \cite{fan_watching_2018_IJCAI}. 

\subsubsection{Point Data}\quad

\wang{Generally}, directly working on all the point data can be heavy and expensive for all the types of consumption mentioned earlier. \wang{Contrarily}, shape \wang{or skeleton} is the main \wang{object} for recognition tasks. Usually, the same shape may be still recognizable with just a subset of points of what we get from the sensors, and in many cases only a small portion of the data are relevant to the recognition task. Therefore, recently quite a few researches have tried to embed data selection components/concepts in their DNN models for handling point data, to achieve a better performance in terms of effectiveness and efficiency.

The selection of point data are usually done by sampling or modeling attention. A simple strategy for sampling is called Furthest Point Sampling (FPS), which can be done efficiently~\cite{PointSel_FPS}, but FPS has significant weaknesses of being task-dependent, low-level (cannot handle semantically high-level representations), permutation-variant, and sensitive
to outliers. Recently, quite a few works have explored new sampling models for point clouds with the help of DNNs. Within them, the work ``Learning to Sample''~\cite{PointSel_S_Net_Dovrat_2019_CVPR} introduces a neural network termed S-NET, which takes a point cloud and produces a smaller one that is optimized for a particular task. The simplified point cloud is not guaranteed to be a subset of the original point cloud, but a post-processing step is adopted to match it to a subset of the original point set. S-NET has a space consumption linearly proportional to its output point set size, and it offers a trade-off between space and inference time. As an example mentioned in the paper, cascading S-NET that samples a point cloud of 1024 to 16 points with a following PointNet~\cite{qi2017pointnet} reduces inference time by over 90\% compared to running PointNet on the complete point cloud, with only 5\% increase in space and 4\% decrease in recognition accuracy (much better than FPS and random sampling). Another more sophisticated model~\cite{PointSel_Yang_2019_CVPR} proposes to do data selection from two perspectives together: a parameter-efficient Group Shuffle Attention (GSA) that does attention based embedding, and an end-to-end learnable and task-agnostic sampling operation called Gumbel Subset Sampling (GSS) for representative subset sampling (from a high-dimention embedding space). Its superiority in effectiveness and efficiency has been shown on several datasets. There are also several other works on attention modeling, such as the Attentional PointNet~\cite{PointSel_Paigwar_2019_CVPR_Workshops} for 3D-Object detection and the Point Attention Network~\cite{PointSel_KingkanOH_BMVC18} for gesture recognition, etc. Due to the space limitation, we don't introduce more details here. Since both the research and applications on point data are now growing very rapidly, \wang{it is expected} that where will be an explosive growth of interests and efforts in designing more efficient and effective recognition models. Data selection for point data are a promising area to look into.

\subsection{Data Representation}

\subsubsection{Image Data}\quad

Image representation for recognition in the DNN era is usually part of the representation learning network, not as a separate pre-recognition operation. However, when there is not enough training data (\emph{i.e.}, small training set) or only a small part of the training data get labeled (\emph{i.e.}, a semi-supervised setting), it can be an effective and also efficient choice to borrow image representation from some suitable pre-trained model (trained on some source data) or from a model trained under an unsupervised setting on the target data before the main task related training with the limited labeled data.

There are some representative references for these two cases. For the first one about borrowing representation from pre-trained models, a typical scenario is medical image analysis or recognition, where \wang{people} have to face the reality of insufficient training data due to many reasons including the privacy issue. A highly cited work in the field \cite{ImageRep_PreTrained_TMI16} discussed about full training vs. fine-tuning (with pre-trained models) for medical imaging applications in three specialties (radiology, cardiology, and gastroenterology) on three vision tasks including classification, detection, and segmentation with extensive experiments, and concluded that the use of a pre-trained CNN with adequate fine-tuning outperformed or, in the worst case, performed as well as a CNN trained from scratch, while at the same time enjoying the benefit of being more robust to the size of training sets (with the help of a layer-wise fine-tuning scheme). A similar conclusion was researched in a study on the effectiveness of using pre-trained CNNs as feature extractors for tuberculosis detection \cite{ImageRep_PreTrained_CBM17}, where three different ways of utilizing pre-trained CNNs (with three different CNN structures) are discussed and in some cases directly using the pre-trained CNNs can even beat their fine-tuned versions. For the second case where data are sufficient but available labels are scarce, pre-training a network under the unsupervised setting, \emph{e.g.} using a spatial prediction task with ``Contrastive Predictive Coding'', has been shown to be effective for fast further training on recognition tasks with little labeled data \cite{ImageRep_Unsupervised_arXiv19}, which is called ``data-efficient'' as it allows task-related training on only a small account of data. Computationally, the task-related training shall be also very efficient, as it can inherit the weights from a pre-trained model which is task independent and can be obtained beforehand. Actually, besides the models trained under unsupervised setting, models trained on a large general dataset (\emph{e.g.} ImageNet) were also found to be very effective for image representation for many visual tasks, superior than the well-designed hand-crafted features \cite{ImageRep_CNN_Features_Baseline_CVPRW14}, and such direct adoption of pre-trained models has been served as a good baseline for exploring new DNN models.

Besides building image representation from a pre-trained model, changing the input image data before training a model from scratch can also be an efficient way. There is interesting finding reported by fast.AI in their MOOC ``Practical Deep Learning for Coder'' and also their write-up\footnote{https://www.fast.ai/2018/08/10/fastai-diu-imagenet/} on how they managed to train Imagenet to 93\% accuracy in just 18 minutes (using 128 NVIDIA V100 GPUs): ``progressive resizing'', namely, using small images for initializing training and then gradually increasing the image size as training progresses, seems to be good for making faster progress on the training and thus saving training time. This raw image presentation strategy seems to be reasonable and effective, and it has been recommended by several other websites, though \wang{there might be still no} any formal academic publication introducing it or justifying it. Further study in this direction can be valuable for both research and applications.

\subsubsection{Video Data}\quad

Data representation for efficient video-based recognition has two representative trends in recent years. Both of them are very new and look rather promising.

One is decomposing video sequences into individual frames for frame-based representation and then represent the motion information by exploring temporal correlations among the high-level features of frames. This is contradictory to the slightly earlier work (Carreira and Zisserman, CVPR 2017 \cite{carreira2017quo}) on the Inception 3D (I3D) architecture which is about 3DCNN. Wu et. al. \cite{Wu_2018T_CN_AAAI} proposed to apply 2DCNN on individual video frames and then do computationally highly efficient 1D temporal convolution on the extracted 2DCNN features, which is both more effective and much more efficient than 3DCNN models for the task of video-based person re-identification. Later, Xie et. al. \cite{Xie_2018_ECCV} replaced the 3D convolutions at the bottom of the 3D CNN network by low-cost 2D convolutions and temporal convolution on the high-level ``semantic'' features (outputs of those 2D convolutions), and also got both better performance and faster speed for action classification. Zhao et. al. \cite{zhao_2018TrajectoryConv_NIPS} further found that making the temporal convolution along the feature trajectories so that the representation can be robust to deformations, and thus they got improved accuracy on action recognition.

The other representative trend is using pooling to generate a super compact and sparse representation for videos before feeding into recognition models. A very successful example in this direction is ``dynamic image'' \cite{bilen_2016DynamicImage_CVPR,bilen_2018DynamicImage_TPAMI}, which is generated by an efficient and effective approximate rank pooling operator, turning a whole video into just one single highly informative image. The dynamic image can simultaneously capture foreground appearance and temporal evolution information, while at the same time excluding irrelevant background appearance information. Therefore, even a simple 2DCNN model built on top of it can generate superior RGB video recognition results. Soon, the model got extended for RGB-D video based activity recognition and showed good results \cite{mukherjee_2018_arXiv}, and very recently it has also been used for generating multi-view dynamic images for the task of depth-video based action recognition \cite{xiao_2019_InfoSci}.

\subsubsection{Point Data}\quad

\begin{figure}
\centering
  {\includegraphics[width=5cm]{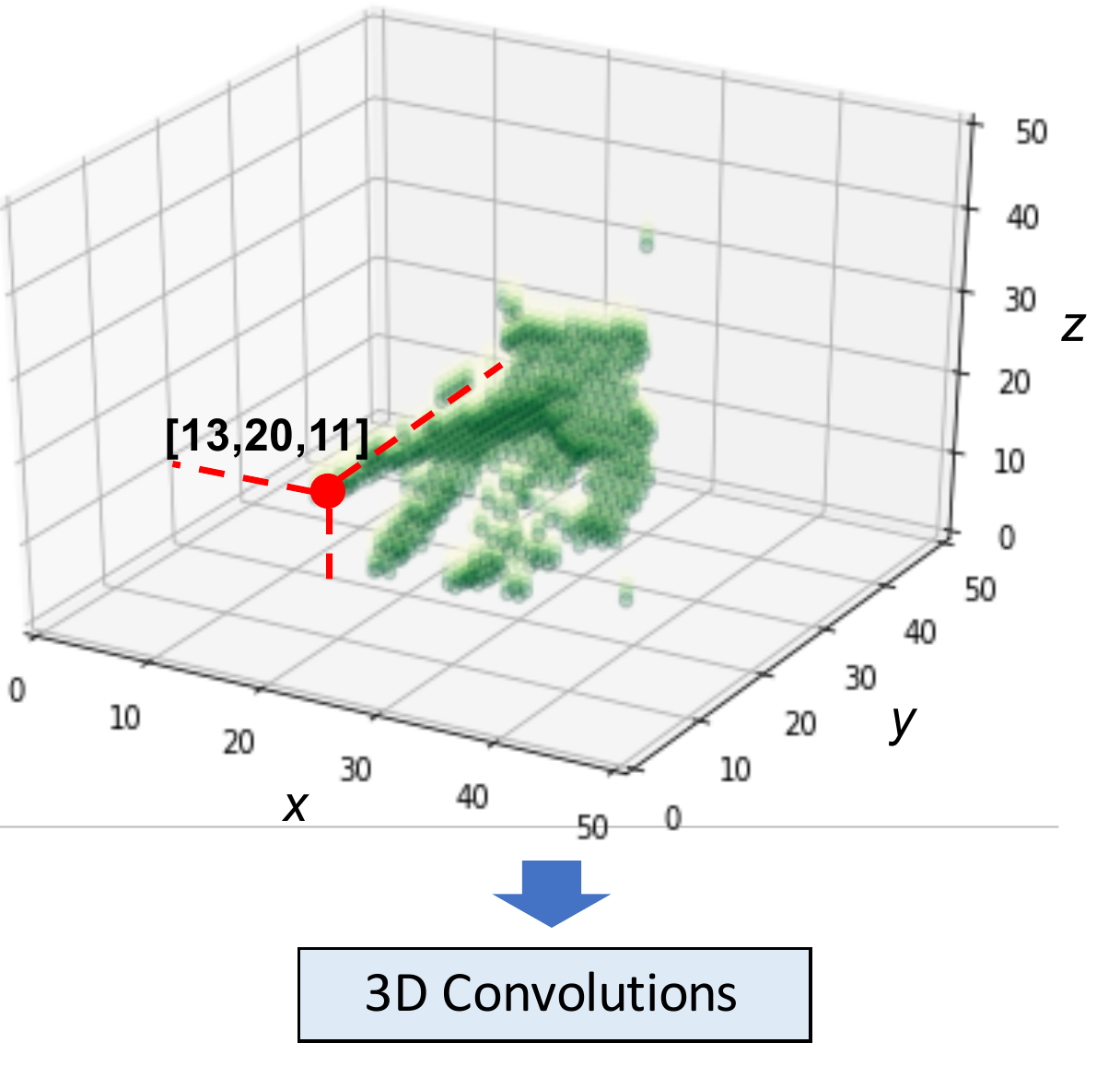}\label{fig_transform}}\\
    \subfloat \footnotesize (a) 3D spatial representation.\\
     \vspace{0.5cm}
 {\includegraphics[width=4.2cm]{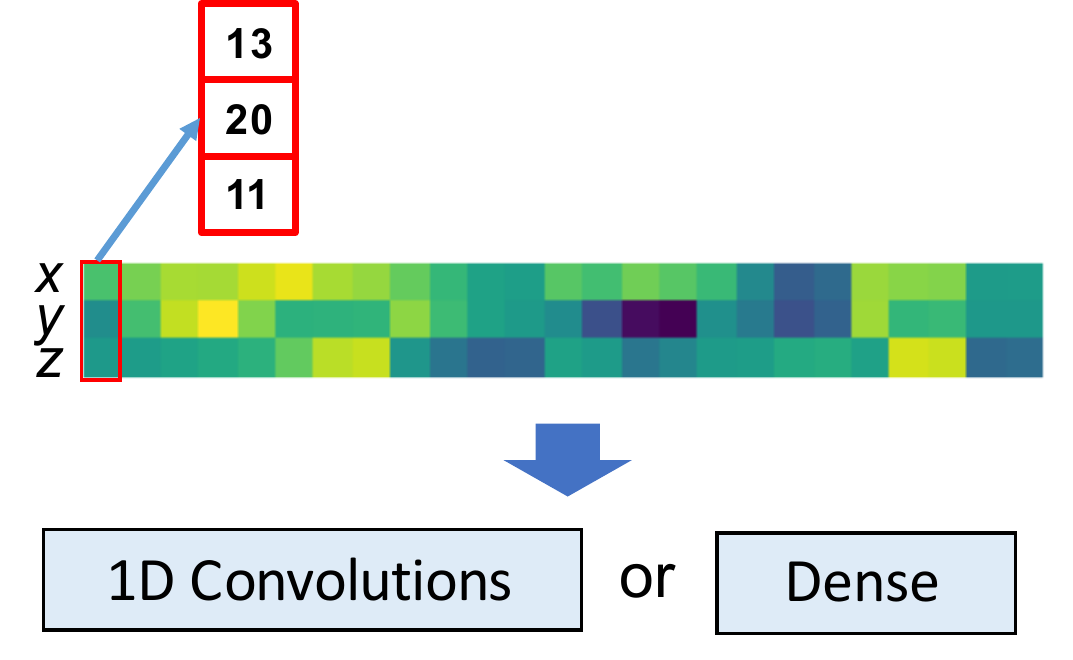}\label{fig_joint_index}}\\
     \subfloat \footnotesize{(b) Flatten coordinates representation. } \\
   \caption{Two kinds of point data representation, and the corresponding neural networks for their processing. }\label{fig:points_represenatations}
\end{figure}

To perform visual recognition tasks on point data, traditional works apply CNN with the same dimension as point data \cite{liu2017two,maturana2015voxnet,moon2018v2v}, which come with huge computation costs, as shown in Fig. \ref{fig:points_represenatations}(a). Compared with RGB data, however, the point data might inherently be sparse. Motivated by such a property, computational-efficient networks are developed. A general approach is to align the coordinates dimension to the CNN channel dimension so that one dimension is reduced compared with the original point data, as shown in Fig. \ref{fig:points_represenatations}(b). Nonetheless, such an approach arises a problem:
the index-adjacent points could be locally uncorrelated in spatial domain no matter how to assign the point indices. Since the RNN/CNN inherently assumes local correlation exists, it is inappropriate to directly process locally uncorrelated features.
Therefore, it is preferred to model all points simultaneously in the network, and let the network automatically learn the proper relationship between them~\cite{yang2019ddnet}.

HCN~\cite{li2018co} and PointNet~\cite{qi2017pointnet} are among the most efficient and superior networks in skeleton-based action recognition and point-cloud-based object recognition, respectively. Although they may look different at the first glance, they all align the coordinates dimension to the CNN channel dimension to make the computation efficient and simultaneously model all points to avoid the influence of point orders. 

\section{On Recognition: Network Compression}
\label{sec:network_compression}

DNNs are always redundant in most cases, thus the great potential ability of compression is inherent in this topic \cite{DenilRedundancy}. Because of some internal high dimensions of raw visual data, corresponding neural networks are redundant consequentially. In this section, \wang{four approaches of network compression in the field of visual recognition are introduced}: 1) compact networks; 2) tensor decomposition; 3) data quantization; and 4) pruning. These approaches can make huge visual recognition networks efficient. Furthermore, some joint compression practices which synthesize multiple specific approaches are also presented.

\subsection{Compact Networks}

\subsubsection{Compact CNNs}\quad

In fact, CNNs are born to deal with visual recognition, and the key natural feature of CNNs is the weights sharing which can be regarded as the earliest practice of compact networks to match the data structure of images. Thus, based on the characteristics of visual recognition tasks, further efforts have been proposed to make CNNs more compact to reduce the ever growing network size \cite{howard2017mobilenets,zhang2018shufflenet,wu2017shift}. In general, most compact design for CNNs can be concluded to two perspectives, one of which is based on the receptive field of filters, and the other one is based on the topology within single convolutional layer (intra-layer) or between convolutional layers (inter-layer). Besides, some more crazy ideas, which invent alternative building blocks for reducing the parameters to learn, appear to be another novel aspect.

\noindent \textbf{Aspect of Receptive Field}. It is clear that designing effective receptive field \cite{luo2016understanding} is crucial to the representation capability of convolutions, which is jointly determined by the filter size and filter pattern. In the aspect of filter size, VGG-Net \cite{simonyan2014very} proposes a stack of two 3 \(\times\) 3 convolutional layers to replace a 5 \(\times\) 5 convolutional layer because their receptive fileds are equivalent. In contrast, more practices are in the aspect of filter pattern. For instance, an n \(\times\) n convolutional layer can be split into two chained layers which have an n \(\times\) 1 layer ahead and a 1 \(\times\) n layer behind \cite{szegedy2016rethinking,paszke2016enet}, atrous or dilated convolution \cite{holschneider1990real,YuKoltun2016,wang2018understanding} uses irregular filters with holes, and deformable convolution \cite{dai2017deformable} generalizes the atrous convolution to learn the offsets of sampling directly from the target tasks.

\begin{figure}
\centering
\includegraphics[width=0.4\textwidth]{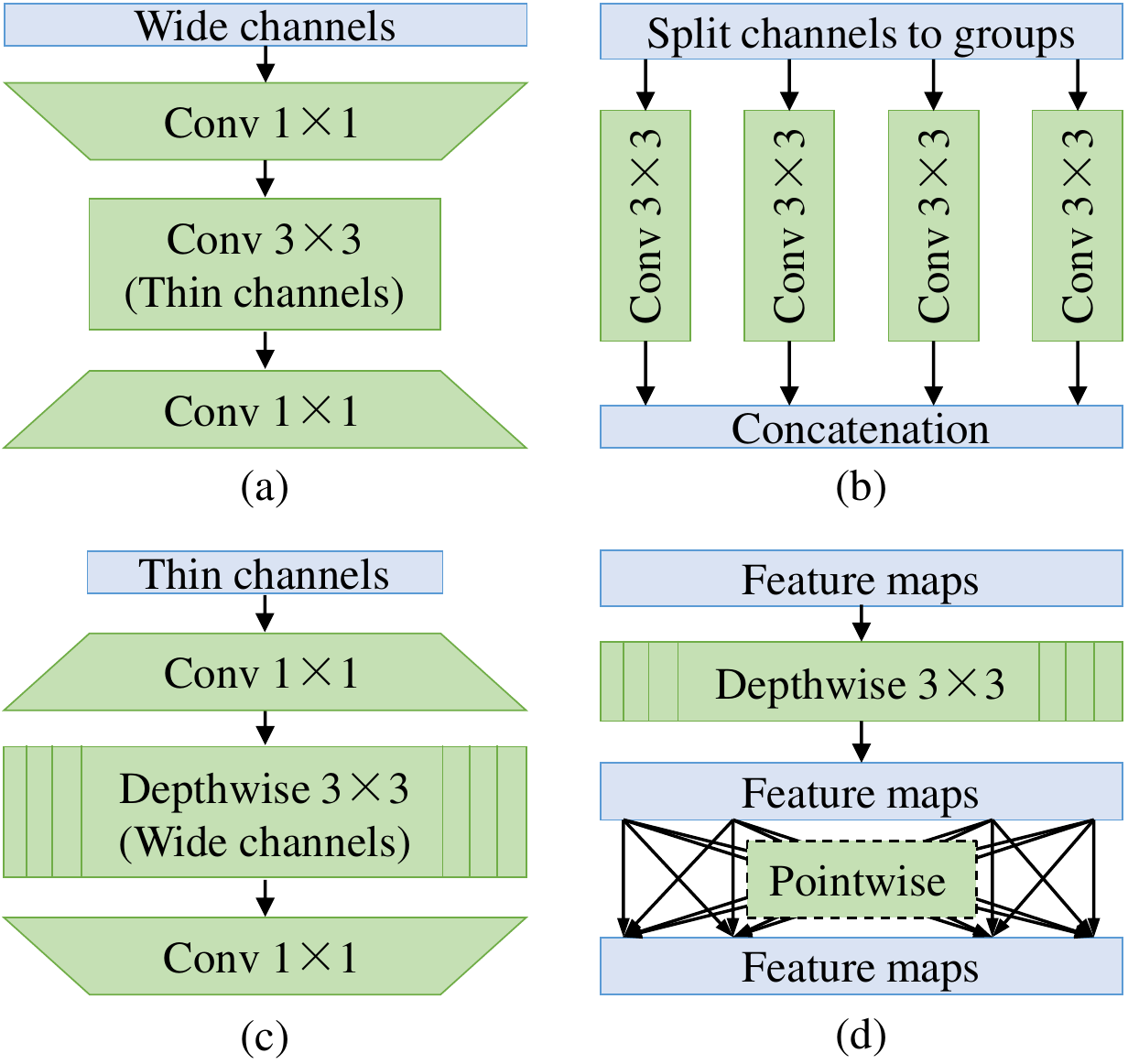}
\caption{Typical efficient inter-layer and intra-layer channel correlations: (a) bottleneck architecture \cite{szegedy2016rethinking}; (b) group convolution with 4 groups \cite{xie2017aggregated}; (c) reversed bottleneck architecture with depthwise convolution \cite{sandler2018mobilenetv2}; (d) depthwise-separable convolution \cite{chollet2017xception}.}
\label{fig_compact_channel}
\end{figure}

\noindent \textbf{Aspect of Topology}. The topology art of changing connections in CNNs can bring more ability of expression or lighter convolutional units. Since the well known NIN \cite{lin2013network} was proposed, it has been aware that vanilla convolutional kernel may be redundant especially for the situation of multiple stacked convolutional kernels. Besides, Inception \cite{Szegedy2015inception} and ResNet \cite{he2016deep,he2016identity} are proposed to enhance the performance of CNNs, which inspired many researchers to think about efficient convolutional units under the well designed topology, \emph{e.g.}, SqueezeNet \cite{Iandola2017squeeze} uses amounts of 1 \(\times\) 1 convolutions to replace 3 \(\times\) 3 convolutions and reduce the counts of channels in the rest 3 \(\times\) 3 convolutions. The residual unit has led to relatively more results of efficient structures, \emph{e.g.}, bottleneck architecture \cite{szegedy2016rethinking} and a similar one with depthwise convolution \cite{chollet2017xception} in ShuffleNet \cite{zhang2018shufflenet}, MobileNetV2 \cite{sandler2018mobilenetv2} and MobileFaceNet \cite{Chen2018mobilefacenet}. Fig. \ref{fig_compact_channel} illustrates several typical compact network designs in the aspect of topology.

\noindent \textbf{New Compact Building Blocks}. The Local Binary Convolutional Neural Networks (LBCNNs) introduced in \cite{CompactNet_LBCNN_Juefei-Xu_2017_CVPR} propose using an alternative of the traditional convolutional layer called local binary convolution (LBC) layer inspired by the design of traditional local binary patterns (LBP), which uses a set of filters with sparse, binary and randomly generated weights that are fixed to replace the traditional convolutional filters whose weights need to be learned. A set (which can be just a small number) of learnable linear weights are used to integrate the feature maps after convolution. Later, a new network type called Perturbative Neural Networks (PNNs) \cite{CompactNet_PNN_juefei-xu2018pnn} replace their each convolutional layer by a so called perturbation layer, which computes its response as a weighted linear combination of non-linearly activated additive noise perturbed inputs. Both the LBC and perturbation layer are shown to be a good approximation of the convolutional layer but with much fewer parameters to learn, and both LBCNNs and PNNs share the same idea of replacing weighted convolution by a linear weighted combination of feature maps.

\subsubsection{Compact RNNs}\quad

RNNs are often used for video recognition to extra the temporal features, but designing a compact RNN is harder than CNN because the inner structures of gated units are complex such as those in LSTM \cite{hochreiter1997long,gers1999learning} and GRU \cite{cho2014learning}. That is to say, any single whole RNN contains several layers in which some units have elaborate internal structures. This situation is the most prominent characteristic which does not exist in neurons of other kinds of DNNs. Therefore, compact RNNs can be designed at the level of units or the level of whole networks. For discussing expediently and concisely, we give a simple equation which can abstract any gated connection, such as forget gate, input gate, output gate, etc.
\begin{equation}\label{CompactBaseRNN}
\bm{y}=\sigma(\bm{W}\bm{x}(t)+\bm{U}\bm{h}(t-1)+\bm{b})
\end{equation}
where \(\bm{y}\) is the data passed the gate; \(\bm{W}\) is the weight matrix corresponding to the input data of current time \(\bm{x}(t)\); \(\bm{U}\) is the weight matrix corresponding to the status of previous time \(\bm{h}(t-1)\); \(\bm{b}\) is the bias vector and the \(\sigma(\cdot)\) is the activation function.

\noindent \textbf{Units Level}.
The topology of gated units in RNNs is complex, however, some approximate reconstruction may be considered to remove some subordinate connections. In fact, GRU is actually a compact design based on LSTM. In detail, there are 4 gated connections described by Equation (\ref{CompactBaseRNN}) in LSTM, whereas GRU has only 3 ones. Thus, space and computation consumption of GRU can be less than that of LSTM. Other than GRU, S-LSTM \cite{wu2016investigating} and JANET \cite{van2018unreasonable} contain forget gates only in their units, minimal gated unit (MGU) \cite{zhou2016minimal} integrates the reset and update gates. On the contrary to the elaborate gates, FastGRNN \cite{Kusupati2019fastgrnn} uses a shared matrix which connects both input and state so that the number of gated connections is kept but parameters are cut down. Quasi-recurrent neural network (QRNN) \cite{Bradbury2017qrnn} replaces the previous output with previous input in the gate calculation, and accordingly gated connections are transformed from Equation (\ref{CompactBaseRNN}) to
\begin{equation}
\bm{y}=\sigma(\bm{W}[\bm{x}(t), \bm{h}(t-1)]+\bm{b}).
\end{equation}

\noindent \textbf{Networks Level}.
One can also simplify the architectures of stacked layers through multiple basic units, just like compact CNNs. Sak et al. \cite{sak2014long} introduce a linear recurrent projection layer to reduce the dimensions of the output of LSTM layers. Wu et al. \cite{wu2016google} introduce residual connections into their 8-layer translation LSTM. Some other analogous compact RNNs include skip-connected RNN \cite{zhang2016architectural}, Grid LSTM \cite{kalchbrenner2015grid} and SRNN \cite{fraccaro2016sequential}.

However, compact design is comparatively hard to implement and has a lack of uniform principles. Even worse, compact method is not easy to achieve high compression ratio, which makes it devoid of significance to deal with large-scaled visual recognition neural networks. Thus, this method is often used to combine with other compression methods or sometimes relied on expensive extra disposing such as the transfer learning based distillation \cite{Hiton2015distill}.

\subsubsection{Neural Architecture Search}\quad

Unlike the rigid human designed or hand-crafted network architectures, \emph{neural architecture search} (NAS) is a promising domain that can automatically construct a compact DNN. For the conventional practice \cite{zoph2017neural}, NAS is the process to train an RNN as the controller to generate a good architecture gradually. In detial, this RNN should train repeatedly to gain the produced architecture, and update the RNN controller itself based on evaluating the architecture by the environment or evaluator. Obviously, this process leads to massive training cost, since the search space will enlarge exponentially when the target DNN is deeper. More broadly, according to \cite{Elsken_2019_NASSurvey}, there are three aspects of challenges lie in the field of NAS, \emph{i.e.}, search space, search algorithm and evaluation strategy, and many researchers aim their attentions to solve these problems. It is interesting that even the controller is usually an RNN, most of the target DNNs are oriented to CNNs, and only a few researches are considered for RNNs \cite{greff2017lstm,zoph_2018_CVPR,liu2018darts,Rawal_2018_Curve}, thus \wang{the review of NAS is mainly focused on CNNs here}.

\begin{table}[!hbp]
\caption{Comparison between state-of-the-art human designed networks (the upper part) and NAS models (the lower part) proposed in the recent years based on ImageNet.}\vspace{3pt}
\label{tab-nas-imagenet}
  \centering
  \renewcommand\arraystretch{1.6}
  \resizebox{0.49\textwidth}{!}{
  \begin{tabular}{c | c c c c }
   \hline
   \hline
    References & \begin{tabular}{c} top-1 Acc \\ (\%) \end{tabular} & \begin{tabular}{c} Params \\ (\(10^6\)) \end{tabular} & \begin{tabular}{c} Ops \\ (\(10^9\)) \end{tabular} & Algo \\
    \hline
    ResNet-152 \cite{he2016deep} & 78.6 & 60.3 & 11.3 & - \\
    DenseNet-264 \cite{huang2017densely} & 79.2 & 32 & 15 & - \\
    ResNeXt-101 \cite{xie2017aggregated} & 80.9 & 44 & 7.8 & - \\
    PolyNet \cite{zhang2017polynet} & 81.3 & 92 & - & - \\
    DPN-131 \cite{chen2017dpn} & 81.5 & 79.5 & 16 & - \\
    \hline
    Hierarchical \cite{liu2017hierarchical} & 79.7 & 64 & - & EA \\
    NAS-Net-A \cite{zoph_2018_CVPR} & 82.7 & 88.9 & 23.8 & RL \\
    PNASNet \cite{liu2018nasrepeating} & 82.9 & 86.1 & 25 & SMBO \\
    AmoebaNet-A \cite{real2018regularized} & 82.8 & 86.7 & 23.1 & EA \\
    TreeCell (mobile) \cite{cai2018path} & 74.6 & - & 0.59 & RL \\
    DARTS (mobile) \cite{liu2018darts} & 73.3 & 4.7 & 0.57 & DGA \\
    Proxy (mobile) \cite{cai2018proxy} & 74.6 & 5.7 & - & \begin{tabular}{c} DGA \\ RL \end{tabular} \\
    EfficientNet \cite{tan2019efficientnet} & 84 & 43 & 19 & RL \\
   \hline
   \hline
  \end{tabular}}
\end{table}

\begin{figure}
\centering
\includegraphics[width=0.85\columnwidth]{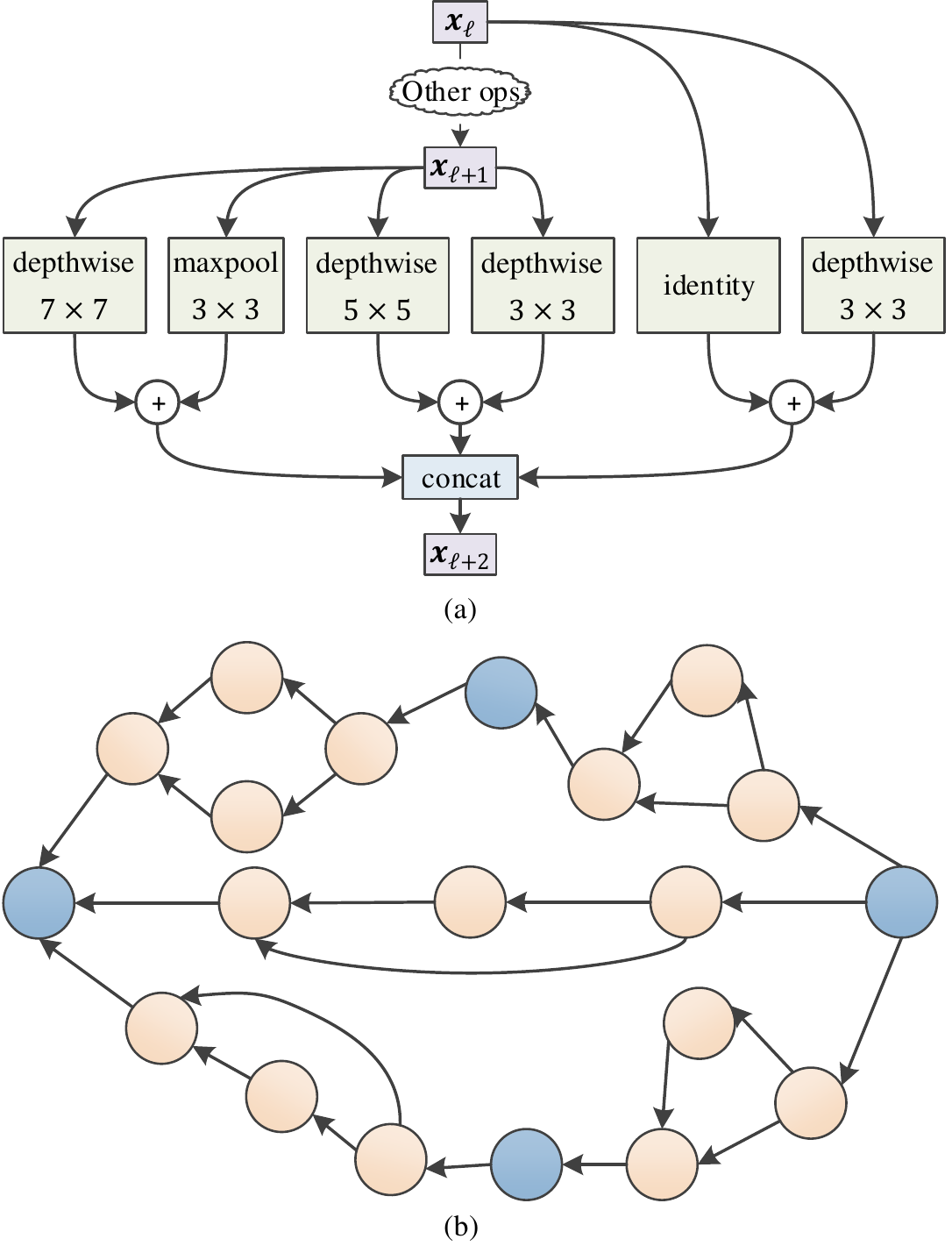}
\caption{Typical cells of NAS: (a) three-block cell \cite{liu2018nasrepeating}; (b) three-level hierarchical cell \cite{liu2017hierarchical}.}
\label{fig_nas_cell}
\end{figure}

\noindent \textbf{Search Space}.
Search space contains various basic network elements, \emph{e.g.}, normal convolution, asymmetric convolution, depthwise convolution, any kind of pooling, different activation functions, etc., and the laws to connect these elements together. Generally, by restricting the search space, the controller can search the basic network cells, which consist of some concrete elements to represent local structures, to construct the target DNN sequentially and and efficiently \cite{zoph_2018_CVPR,Zhong_2018_CVPR}. Moreover, the search complexity can be further reduced in the range of the cell, \emph{e.g.}, progressively increasing the number of blocks within one layer as shown in Figure \ref{fig_nas_cell}(a), and searching hierarchical topology within a cell as illustrated in Figure \ref{fig_nas_cell}(b).

\noindent \textbf{Search Algorithm}.
Distinctly, appropriate search algorithm is critical since its target is to efficiently combine basic network elements together to be a whole network that fits a specific task or dataset most. Roughly speaking, except naive random search, these algorithms can be classified into four approaches, \emph{i.e.}, a) reinforcement learning (RL) including Q-learning \cite{baker2017designing,zhong2017practical}, REINFORCE \cite{zoph2017neural,cai2018efficient,cai2018path} and proximal policy optimization (PPO) \cite{zoph_2018_CVPR}; b) neural evolutionary algorithm (EA) \cite{real2018regularized,liu2017hierarchical,xie2017genetic}; c) Bayesian optimization (BO) \cite{Kandasamy_NASBO,Klein_NASBO}; and d) differentiable gradient for architecture (DGA) \cite{cai2018proxy,liu2018darts}. It is hard to say which algorithm is better, however, classical RL approach still reveals the higher boundary of performance \cite{tan2019efficientnet}. Nowadays, some other minor improvements are also used for improving the searching efficiency, \emph{e.g.}, exploring the search space with the network transformation and weight reusing \cite{cai2018efficient,cai2018path,Elsken_2019_MONAS}, and sequential model based optimization (SMBO) \cite{liu2018nasrepeating} which can choose the more likely direction of optimization rather than blindly searching with wide range.

\noindent \textbf{Evaluation Strategy}.
It is clear that the searched architecture should be evaluated to verify whether it can fit the task. The first strategy is simple, \emph{i.e.}, finitely train and evaluate the target DNN to observe its trends, \emph{e.g.}, just train on subset of dataset \cite{Klein_Fast}, build an accuracy predictor trained on the limited search space to guide the following search process \cite{liu2018nasrepeating,Cai_Fast}, predict the search direction based on partial training curves of current searched architectures \cite{Klein_2017_Curve,Rawal_2018_Curve}, etc. The second way is to yield the searched architecture to inherit the weights which are produced in previous searched architectures \cite{Wei_2016_Morphism,cai2018efficient,cai2018path}. The third aspect is termed as one-shot NAS, that only train the one-shot model, from which the searched architectures can directly share the one-shot weights without extra training \cite{liu2018darts,cai2018proxy}.

To fully show its promising developing tendency, we make Table \ref{tab-nas-imagenet} to compare some well-known human designed networks with NAS models nowadays. On the whole, NAS models can surpass human designs in both accuracy and efficiency, especially the EfficientNet \cite{tan2019efficientnet} with the highest accuracy and several lightweight models for mobile format \cite{cai2018path,liu2018darts,tan2019efficientnet} with tiny number of parameters and operations (GFLOPs). Further, the searching time of NAS algorithms has also decreased significantly from 22,400 \cite{zoph2017neural} to 4 \cite{liu2018darts} GPU days, that makes it possible to apply NAS models on the real embedded devices. Nevertheless, there are still some problems to be solved in this direction, \emph{e.g.}, a lack of practical report of deploying NAS models in fickle real data, enhancing the NAS models for mobile platforms to find the limitation of compact architecture, handling the trade off between accuracy and size, etc. Hence, \wang{all the signs indicate that} this kind of approach is promising but still needs further studies.

\subsection{Tensor Decomposition}

Traditionally, since the parameters in DNNs are mostly stored as matrices, matrix decomposition is widely used in network compression, especially the SVD \cite{MasanaSVD,KumamotoInputSVD,DebInputSVD,ZouSVDNet,ZhangSVDSt,ZhangSVDFilterGroup,IoannouSVDFilterGroup,PengSVDFilterGroup}. However, restriction of orders makes it limited to use matrix decomposition to compress larger and larger visual recognition neural networks. Besides, tensors may have some inherent connections with neural networks \cite{HuTensorDNN}. Hence, \wang{discussing tensor decomposition here is sufficient} because a matrix is actually a 2nd-order tensor.

\subsubsection{Classical Tucker Decomposition}\quad

No matter CP decomposition \cite{CarollCP} or HOSVD \cite{DeLathauwerHOSVD1,DeLathauwerHOSVD2}, all classical tensor decomposition methods for a \(d\)th-order tensor \(\bm{\mathcal{A}} \in \mathbb{R} ^{n_{1} \times n_{2} \times \cdots \times n_{d}}\) can be represented uniformly in Tucker decomposition \cite{TuckerTucker} like \cite{KoldaTensorOp}
\begin{equation}\label{ClassicalTensorDecomposition}
\bm{\mathcal{A}} = \bm{\mathcal{K}} \times_{1} \bm{F}^{(1)} \times_{2} \bm{F}^{(2)} \times_{3} \cdots \times_{d} \bm{F}^{(d)}.
\end{equation}
where \(\bm{\mathcal{K}} \in \mathbb{R} ^{r_{1} \times r_{2} \times \cdots \times r_{d}}\) denotes the kernel tensor, any one \(\bm{F} ^{(i)} \in \mathbb{R} ^{r_{i} \times n_{i}}\) (\(i \in \{ 1,2,\cdots,d \}\)) is the factor matrix, and the operation \(\times_{i}\) means the mode-\( i \) contracted product \cite{KoldaTensorOp}. If every \( r_{i} \) equals to a positive integer \( r_{C} \) and the kernel tensor \( \bm{ \mathcal{K} } \) presents like a superdiagonal tensor, which means all elements in \( \bm{ \mathcal{K} } \) are 0 except \( \mathcal{K} (x_{1},x_{2},\cdots,x_{d}) \) with \( x_{1}=x_{2}=\cdots=x_{d} \), then Equation (\ref{ClassicalTensorDecomposition}) will become CP (Canonical Polyadic) decomposition \cite{KoldaTensorOp}. If all the factor matrices \( \bm{F} ^{(i)} \) are orthogonal and the kernel tensor \( \bm{ \mathcal{K} } \) is so called all-orthogonal \cite{DeLathauwerHOSVD1,DeLathauwerHOSVD2}, then Equation (\ref{ClassicalTensorDecomposition}) will become HOSVD.

There are many researchers who apply CP and Tucker to compress the weights in neural networks in recent years, especially CNNs for visual recognition \cite{KimTucker-2,ChienTuckerLayer,KossaifiTuckerTCL1,KossaifiTuckerTCL2,JanzaminCPBP,LebedevCPTune,AstridCPCNN,TranCPTucker,SchuttTensorNN,ZhouNNToLearnCPRank,OymakCP}. In general, to utilize Equation (\ref{ClassicalTensorDecomposition}) efficiently, any single weight matrix \( \bm{W} \in \mathbb{R} ^{M \times N} \) should be mapped into a \(d\)th-order tensor \( \bm{ \mathcal{W} } \in \mathbb{R} ^{m_{1}n_{1} \times m_{2}n_{2} \times \cdots \times m_{d}n_{d}} \) where \(M = \prod_{i=1}^d m_i\) and \(N = \prod_{i=1}^d n_i\) \cite{NovikovTT}. Thus, the greater value of \(d\) can bring the more effective compression which decreases the complexity from \( \mathcal{O} ((mn) ^{d}) \) to \( \mathcal{O} (dmnr + r ^{d}) \). However, the curse of dimensionality \cite{GrasedyckTensorSurvey,CichockiTensorApp} has not been solved completely by Tucker because of \(\mathcal{O}(r^{d})\). In addition, the relatively new block term decomposition (BTD) \cite{DeLathauwerTuckerBTD}, which is in actual the sum of multiple Tucker blocks, can ease this curse to some extent, since the size of each kernel tensor may be smaller. In \cite{Ye_2018_BTD}, BTD-LSTM shows superior ability of keeping information for visual recognition, but corresponding computation process is relatively complex.

\subsubsection{Tensor Network}\quad

\begin{table*}
\caption{Applications of neural networks compressed by tensor networks for visual recognition tasks.}\vspace{3pt}
\label{Table-TT-Application}
  \centering
  \renewcommand\arraystretch{1.6}
  \resizebox{0.95\textwidth}{!}{
  \begin{tabular}{c | c c c c c}
   \hline
   \hline
     References & Format & Compressed Parts & Dataset & Compression Ratio & Accuracy Loss \\
     \hline
    Novikov et al. (2015) \cite{NovikovTT} & TT CNN & FC & CIFAR10 & 11.9\(\times\) & 1.26\% \\
    Zhao et al. (2018) \cite{ZhaoTR2} & TC CNN & FC & CIFAR10 & 444\(\times\) / 1300\(\times\) & 0.13\% / 2.18\% \\
    Huang et al. (2018) \cite{HuangTTCNN} & TT CNN & FC & MINST & 14.85\(\times\) & 1.5\% \\
    Su et al. (2018) \cite{SuTTCNN} & TT CNN & FC & MINST & 500\(\times\) & 2\% \\
    Garipov et al. (2016) \cite{GaripovTTCNN} & TT CNN & Conv \& FC & CIFAR10 & 82.87\(\times\) & 1.1\% \\
    Wang et al. (2020) \cite{Wang_2020_TT3DCNN} & TT 3DCNN & Conv \& FC & UCF11 / ModelNet40 & 107.5\(\times\) / 160.7\(\times\) & 0.22\% / 0.14\% \\
    Tjandra et al. (2017) \cite{TjandraTTRNN1} & TT GRU & $\pmb{W}$ \& $\pmb{U}$ & Sequential MNIST & 43.52\(\times\) / 69.80\(\times\) & 0.3\% / 0.3\% \\
    Yang et al. (2017) \cite{YangTTRNN} & \begin{tabular}{c} TT LSTM \\ TT GRU \end{tabular} & $\pmb{W}$ & UCF11 / Hollywood2 & \begin{tabular}{c} 17554.3\(\times\) / 23158.8\(\times\) \\ 13687.1\(\times\) / 18313\(\times\) \end{tabular} & \begin{tabular}{c} -30.4\% / -43.8\% \\ -32.5\% / -28.8\% \end{tabular} \\
    Pan et al. (2018) \cite{PanTCRNN} & TC LSTM & $\pmb{W}$ & UCF11 / HMDB51 & 34192\(\times\) & -1.5\% / -0.9\% \\
    Wu et al. (2020) \cite{Wu_2020_Hybrid} & HT LSTM & $\pmb{W}$ \& $\pmb{U}$ & UCF 11 / UCF 50 & 58.41\(\times\) / 57.96\(\times\) & 0.12\% / 1.37\% \\
    Yin et al. (2020) \cite{Yin_2020_HTRNN} & HT LSTM & $\pmb{W}$ & UCF 11 / Youtube Face & 47375\(\times\) / 72818\(\times\) & -17.5\% / -54.9\% \\
   \hline
   \hline
  \end{tabular}}
  \vspace{-5pt}
\end{table*}

Tensor network \cite{CichockiTensorNetworks,PellioniszTensorNetwork}, which represents a tensor with a linked of matrices or low order tensors with contracted products, is promising to avoid the curse of dimensionality by eliminating the high order kernel tensor with \( r ^{d} \) elements according to Equation (\ref{ClassicalTensorDecomposition}). Commonly, there are three types of tensor networks are applied, \emph{i.e.}, tensor train (TT) \cite{OseledetsTTInvent1,OseledetsTTInvent2}, tensor chain (TC) \cite{KhoromskijQTT,EspigTC} or tensor ring (TR) \cite{ZhaoTR1,ZhaoTR2}, and hierarchical Tucker (HT) \cite{HackbuschInventHT,GrasedyckInventHT}.

\noindent \textbf{Tensor Train}.
According to \cite{LeeHTTT}, a \( d \)th-order tensor \( \bm{\mathcal{A}} \in \mathbb{R} ^{n_{1} \times n_{2} \times \cdots \times n_{d}} \) can be represented as
\begin{equation}\label{TTBaseTensor}
\bm{\mathcal{A}} = \bm{\mathcal{G}}_{1} \times ^{1} \bm{\mathcal{G}}_{2} \times ^{1} \cdots \times ^{1} \bm{\mathcal{G}}_{d}
\end{equation}
where the operation \( \times ^{1} \) is the mode-\( (N,1) \) contracted product \cite{LeeTensorOp}, and the core tensors \( \bm{\mathcal{G}}_{i} \in \mathbb{R} ^{r_{i-1} \times n_{i} \times r_{i}} \) (\(i={1,2,\cdots,d}\)) always have \( r_{0}=r_{d}=1 \). Apparently, according to the equation above, the spatial complexity of TT format is \( \mathcal{O}(dnr^{2}) \) where \( n \) is the maximum value of modes and \( r \) is the maximum value of TT ranks. It is obvious that the \(\mathcal{O}(r^{d})\) in classical decomposition, \emph{i.e.}, Equation (\ref{ClassicalTensorDecomposition}), is avoided and the curse of dimensionality is solved.

\begin{figure}
\centering
\includegraphics[width=0.45\textwidth]{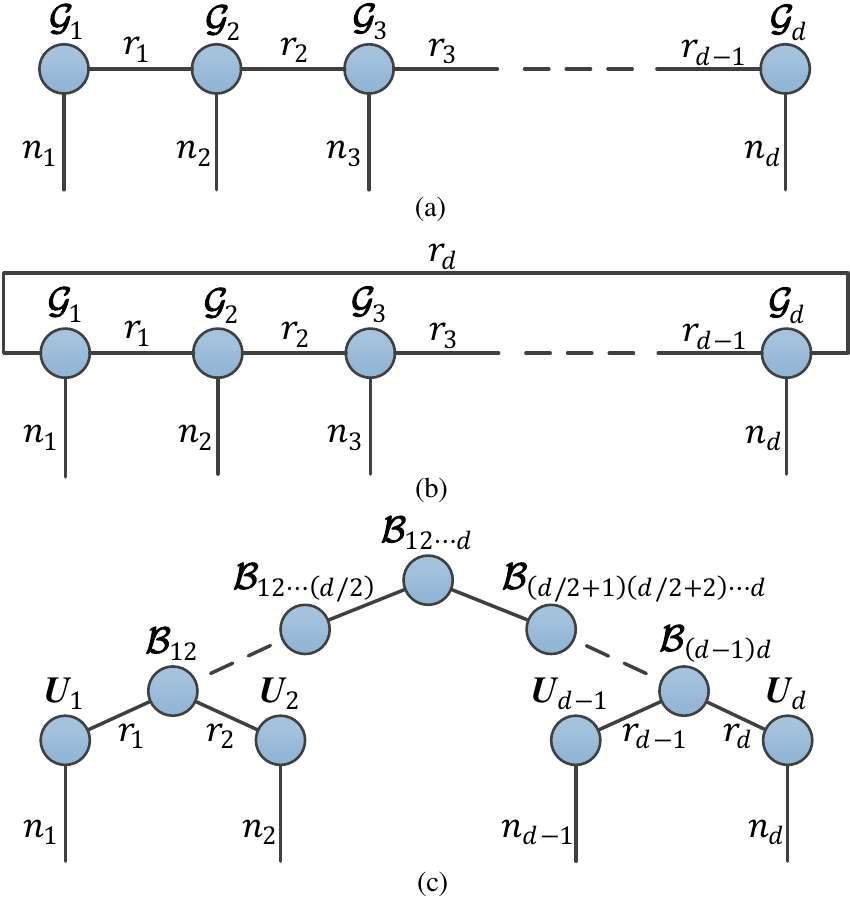}
\caption{Tensor network graphs for (a) tensor train, (b) tensor chain, and (c) hierarchical Tucker format of  a \( d \)th-order tensor \( \bm{\mathcal{A}} \in \mathbb{R} ^{n_{1} \times n_{2} \times \cdots \times n_{d}}\).}
\label{fig_tn}
\end{figure}

\noindent \textbf{Tensor Chain}.
TC format, which can be regarded as a variant of the TT format, is first proposed by B. N. Khoromskij \cite{KhoromskijQTT} and proved having similar approximation capability as TT by Espig et al. \cite{EspigTC}. Nowadays, the TC format is introduced in the field of DNNs as the TR format by Zhao et al. \cite{ZhaoTR1,ZhaoTR2}. Comparing with TT, the TC format of a \( d \)th-order tensor \( \bm{\mathcal{A}} \in \mathbb{R} ^{n_{1} \times n_{2} \times \cdots \times n_{d}} \) has only one difference which is \( r_{0} = r_{d} \neq 1 \). Thus, differ from Equation (\ref{TTBaseTensor}), there are two pairs of equal modes must be contracted at the end like
\begin{equation}\label{TCBaseTensor}
\bm{\mathcal{A}} = (\bm{\mathcal{G}}_{1} \times^{1} \bm{\mathcal{G}}_{2} \times^{1} \cdots \times^{1} \bm{\mathcal{G}}_{d-1}) \times_{1,d+1}^{3,1} \bm{\mathcal{G}}_{d}
\end{equation}
where \( (\bm{\mathcal{G}}_{1} \times^{1} \bm{\mathcal{G}}_{2} \times^{1} \cdots \times^{1} \bm{\mathcal{G}}_{d-1}) \in \mathbb{R} ^{r_{d} \times n_{1} \times n_{2} \times \cdots \times n_{d-1} \times r_{d-1}} \), \( \bm{\mathcal{G}}_{d} \in \mathbb{R} ^{r_{d-1} \times n_{d} \times r_{d}} \), and \(\times_{1,d+1}^{3,1}\) means the paired contracted modes are, the 1st of the former vs. the 3rd of the latter while the (\(d+1\))th of the former vs. the 1st of the latter. Fig. \ref{fig_tn} shows TT and TC in tensor network graphs, where every node represents a tensor and each edge is a mode of its connected tensor.

\noindent \textbf{Hierarchical Tucker}.
In fact HT is the fountainhead of TT, \emph{i.e.}, TT is a special form of HT \cite{CohenHT}, since HT has extremely flexible organizational structure. Particularly, for a \(d\)th-order tensor \( \bm{\mathcal{A}} \in \mathbb{R} ^{n_{1} \times n_{2} \times \cdots \times n_{d}} \), \wang{their modes could be divided} into two sets as $t=\{ {t_1},{t_2},...,{t_k}\}$ and $s=\{ {s_1},{s_2},...,{s_{d - k}}\}$, then we have
\begin{equation}\label{HT_Base}
\bm{U}_{t} = (\bm{U}_{t_l} \otimes \bm{U}_{t_v})\bm{B}_{t}
\end{equation}
where $\bm{U}_{t}\in\mathbb{R}^{{n_{{t_1}}}{n_{{t_2}}}...{n_{{t_k}}} \times{r_t}}$, $\bm{U}_{t_l}\in\mathbb{R}^{{n_{{t_{l_1}}}}{n_{{t_{l_2}}}}...{n_{{t_{l_i}}}} \times{r_{t_l}}}$, $\bm{U}_{t_v}\in\mathbb{R}^{{n_{{t_{v_1}}}}{n_{{t_{v_2}}}}...{n_{{t_{v_{k-i}}}}} \times{r_{t_v}}}$ are called truncated matrices, $\bm{B}_{t}\in\mathbb{R}^{{{r_{{t_l}}}{r_{{t_v}}}} \times{r_t}}$ is termed as transfer matrix, and \(\otimes\) is Kronecker product. One can continuously using Equation (\ref{HT_Base}) until all the truncated matrices become \(\bm{U}_{i} \in \mathbb{R} ^{n_i \times r_i}\), thus the HT format of \(\bm{\mathcal{A}}\) will be done. Obviously, there are multiple ways to split the modes of \(\bm{\mathcal{A}}\), however, even the simplest format with the binary tree is still formidable to write in formulation \cite{Yin_2020_HTRNN,Wu_2020_Hybrid}. Fortunately, tensor network graph in Fig. \ref{fig_tn}(c) can afford us a convenient description.

\subsubsection{Typical Practices}\quad

On the whole, TT format represents to be the most vibrant method of tensor decomposition. We list current applications of TT compressed neural networks including both CNNs and RNNs for visual recognition in Table \ref{Table-TT-Application}, where a handful of TC and HT are introduced as well. We find an interesting phenomenon that the tensor decomposed CNNs for image tasks are hard to avoid the accuracy loss, while RNNs for video tasks are easy to achieve higher accuracy in compressed forms. We guess that RNNs have larger scale than CNNs in general which situation verifies that a larger network is easier to compress \cite{zhu2017prune}, and the practice on 3DCNNs further verified  this since 3D convolutional kernels have heavier redundancy \cite{Wang_2020_TT3DCNN}. However, how to deal with the accuracy loss still needs to be learnt.

\subsection{Data Quantization}
\label{subsection:quantization}

\begin{figure*}
\centering
\includegraphics[width=0.9\linewidth]{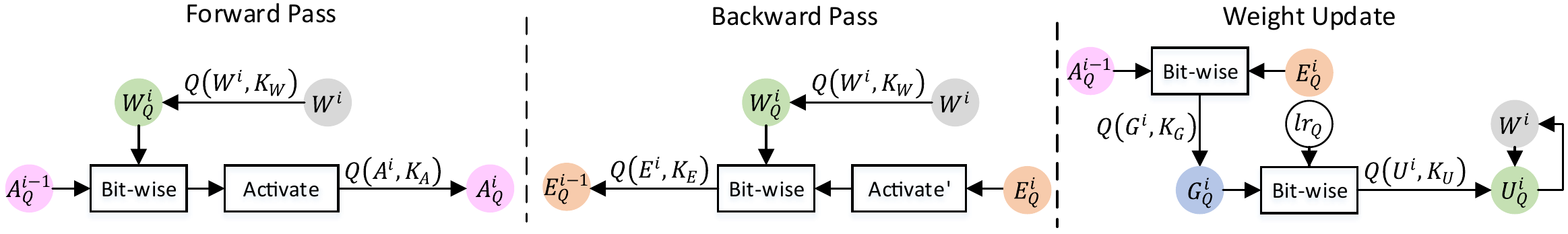}
\caption{Data of quantized objects including \(W_Q\), \(A_Q\), \(G_Q\), \(E_Q\) and \(U_Q\).}
\label{fig_q_object}
\end{figure*}

Data quantization can project concrete data, including weight matrices, gradients, nonlinear activation functions, etc., from high-precision value space to low-precision value space, \emph{e.g.}, from real number field \(\mathbb{R}\) to integer field \(\mathbb{N}\). This approach can not only reduce the space complexity of network parameters, but also accelerate the running time of neural networks because bit operations are much faster than float operations. As raw visual data are generally represented by integers, quantized neural networks may be more suitable for visual recognition tasks. 

\subsubsection{Method}\quad

\noindent \textbf{Problem Formulation}.
In general, there are two formulations to describe how to transform floating point weights, gradients, or activations, etc., to quantized data type. One category, which is the most widely used, projects the original high-precision value to the space of quantized data as
\begin{equation}
Q(x)=\Delta \cdot round(\frac{x}{\Delta})
\label{Quan_k}
\end{equation}
where $x$ is the original high-precision value in the continuous space, \(Q(x)\) is the quantized data in a discrete space, $round(\cdot)$ is the rounding operation, and $\Delta$ is the quantization step length if the discrete states have uniform distribution. If $K$-bit quantization is used, we have $\Delta=\frac{1}{2^{K-1}}$ to discretize $x\in [0,~1]$ to $2^K$ states. This category, as described in Equation (\ref{Quan_k}), is straightforward and easy to consider, thus most practices have followed this direction, which can be observed in Table \ref{Table-quantization}.

The other category regards the quantization as an optimization problem and tries to solve it approximately. The classic model can be generally governed as
\begin{equation}
\underset{Q}{\text{min}} ~\|\pmb{X} - Q(\pmb{X})\|_2^2,~s.t.  ~Q_i\in X_Q~for~all~i
\label{Quan_Optimization}
\end{equation}
where $X_Q=\{Q_1, Q_2, \cdots, Q_n\}(i \in \{1,2,\cdots,n\})$ is a set which has \(n\) discrete states for quantization. The earliest and the most well-known quantization in the category of optimization is the XNOR-NET \cite{rastegari2016xnor}. An obvious motivation is that the optimization problem pays more attention to the whole network rather than local quantization, so Equation (\ref{Quan_Optimization}) may be more appropriate for large scale neural networks.

\noindent \textbf{Quantized Objects}.
As mentioned above, except weight (W), there are also several other different objects in neural networks can be quantized, such as activation (A), error (E), gradient (G), and weight update (U). Fig. \ref{fig_q_object} illustrates the data of these quantized objects \(W_Q\), \(A_Q\), \(G_Q\), \(E_Q\) and \(U_Q\) existing in forward pass, backward pass and weight update processes. Parameter (W) is the most straightforward to be dealt with. Propagation data (A,E) correlate highly to the data flow during forward and backward passes which influence accelerating a lot. Quantized gradient and update have a great help for training the full quantized networks, but are harder to implement, thus the corresponding practices are fewer as shown in Table \ref{Table-quantization}.

\noindent \textbf{Algorithm Description}.
Generally, if the bit-width \(K\) is given, Equation (\ref{Quan_k}) can be rewritten as
\begin{equation}
x_Q = Q(x,K)
\label{Quan_output}
\end{equation}
where \(x_Q\) is the quantized data with \(K\)-bits, \emph{i.e.}, quantized objects \(W_Q\), \(A_Q\), \(G_Q\), \(E_Q\) and \(U_Q\) described above. Data flow among these quantized data can generally be computed by bit-wise operation which is much faster than the full precision computation. Besides, performance of quantized networks with appropriate bit-width will not degenerate that can be learnt in Table \ref{Table-quantization}. The overall algorithm is briefly described in Fig. \ref{fig_q_object}, and the more comprehensive algorithm description of quantization can consult \cite{Yang_2019_WAGEU}.

\begin{table*}
\caption{Typical practices of quantization for CNNs.}\vspace{3pt}
\label{Table-quantization}
  \centering
  \renewcommand\arraystretch{1.6}
  \resizebox{0.95\textwidth}{!}{
  \begin{tabular}{c | c c c c c}
   \hline
   \hline
     References & Formulation & Objects & State Distribution & State Projection & Performance \\
     \hline
     VQN (2018) \cite{achterhold2018variational} & Optimization & W & Uniform & Deterministic & CIFAR10, DenseNet, 91.22\% \\
     ADMM (2017) \cite{leng2017extremely} & Optimization & W & Non-uniform & Deterministic & ImageNet, ResNet50, 72.5\% \\
     BinaryRelax (2018) \cite{yin2018binaryrelax} & Optimization & W & Uniform & Deterministic & ImageNet, ResNet18, 66.5\% \\
     Sketching (2017) \cite{guo2017network} & Optimzation & W & Non-uniform & Deterministic & ImageNet, AlexNet, 55.2\% \\
     INQ (2017) \cite{zhou2017incremental} & Projection & W & Non-uniform & Deterministic & ImageNet, ResNet18, 66.02\% \\
     HWGQ (2017) \cite{Cai2017Deep} & Projection & W,A & \begin{tabular}{c} Uniform \\ Non-uniform \end{tabular} & Deterministic & ImageNet, ResNet50, 64.6\% \\
     TBN (2018) \cite{wan2018tbn} & Optimization & W,A & Uniform & Deterministic & ImageNet, ResNet34, 58.2\% \\
     LQ-Net (2018) \cite{zhang2018lq} & Optimization & W,A & Non-uniform & Deterministic & ImageNet, ResNet50, 71.5\% \\
     Balanced DoReFa (2017) \cite{zhou2017balanced} & Projection & W,A & Uniform & Deterministic & ImageNet, ResNet18, 59.4\% \\
     WRPN (2017) \cite{mishra2017wrpn} & Projection & W,A & Uniform & Deterministic & ImageNet, ResNet32, 73.32\% \\
     Regularization (2018) \cite{choi2018learning} & Projection & W,A & Uniform & Deterministic & ImageNet, ResNet18, 61.7\% \\
     BWNH (2018) \cite{Hu2018Hashing} & Optimization & W & Non-uniform & Stochastic & ImageNet, ResNet18, 64.3\% \\
     Group-Net (2018) \cite{zhuang2018training} & Projection & W,A & Uniform & Deterministic & ImageNet, ResNet50, 73.4\% \\
     UNIQ (2018) \cite{baskin2018uniq} & Projection & W,A & Non-uniform & Deterministic & ImageNet, ResNet32, 72.6\% \\
     FAQ (2018) \cite{mckinstry2018discovering} & Projection & W,A & Uniform & Deterministic & ImageNet, ResNet152, 78.54\% \\
     AdaptiveQ (2018) \cite{zhou2018AdaptiveQ} & Projection & W & Uniform & Deterministic & ImageNet, ResNet50, 75\% \\
     LWQ (2018) \cite{zhu2018AdaptiveLWQ} & Optimization & W,A & Uniform & Stochastic & ImageNet, AlexNet, 45.64\% \\
     HAQ (2019) \cite{wang2019haq} & Projection & W,A & Uniform & Deterministic & ImageNet, ResNet50, 76.14\% \\
     TernGrad (2017) \cite{wen2017terngrad} & Projection & G & Uniform & Stochastic & ImageNet, AlexNet, 57.61\% \\
     TQN (2017) \cite{li2017training} & Projection & W,U & Uniform & \begin{tabular}{c} Deterministic \\ Stochastic \end{tabular} & ImageNet, ResNet18, 47.89\% \\
     DST (2018) \cite{li2018training} & Projection & W,U & Uniform & Stochastic & CIFAR10, VGG8, 88.16\% \\
     GXNOR-Net (2018) \cite{deng2018gxnor} & Projection & W,A,U & Uniform & \begin{tabular}{c} Stochastic (W) \\ Deterministic (A) \end{tabular} & CIFAR10, VGG8, 92.5\% \\
     MP (2017) \cite{micikevicius2017mixed} & Projection, & W,A,E & Uniform & Deterministic & ImageNet, ResNet50, 76.04\% \\
     MP-INT (2018) \cite{das2018mixed} & Projection, & W,A,E & Uniform & Deterministic & ImageNet, ResNet50, 75.77\% \\
     QBPv2 (2018) \cite{banner2018scalable} & Projection, & W,A,E & Uniform & Deterministic & CIFAR10, ResNet18, 89.2\% \\
     Flexpoint (2017) \cite{koster2017flexpoint} & Projection, & W,A,G & Uniform & Deterministic & CIFAR10, ResNet110, \(>\) 94\% \\
     WAGE (2018) \cite{Wu_2018_WAGE} & Projection, & W,A,G,E,U & Uniform & \begin{tabular}{c} Deterministic (W,A,E,U) \\ Stochastic (G) \end{tabular} & ImageNet, AlexNet, 48.4\% \\
     FX Training (2018) \cite{Sakr_2019pertensor_ICLR} & Projection, & W,A,G,E,U & Uniform & Deterministic & CIFAR10, ResNet20, 92.76\% \\
     8b Training (2018) \cite{wang2018training} & Projection, & W,A,G,E,U & Uniform & \begin{tabular}{c} Deterministic \\ Stochastic \end{tabular} & ImageNet, ResNet50, 71.72\% \\
     OCS (2019) \cite{Zhao_2019_OCS} & Projection & W,A & Non-uniform & Stochastic & ImageNet, ResNet50, 75.7\% \\
     Full 8-bit (2020) \cite{Yang_2019_WAGEU} & Projection & W,A,G,E,U & Uniform & Deterministic & ImageNet, ResNet50, 69.07\% \\
     AutoQ (2020) \cite{Lou_2020_AutoQ} & Optimization & W,A & Non-uniform & Stochastic & ImageNet, ResNet50, 74.47\% \\
   \hline
   \hline
  \end{tabular}}
  \vspace{-5pt}
\end{table*}

\noindent \textbf{State Distribution and Projection}.
The data in quantization set always have concrete discrete states, which may contain different distributions. For example, uniform distribution is the most widely used one, logarithmic distribution has exponential variance on step length that has obvious benefits to convert multiplication to addition, and adaptive distribution often occurs in the situation when formulating the quantization as an optimization problem such as TTQ \cite{zhu2016trained}, ADMM \cite{leng2017extremely}, etc. On the other hand, the primary mission in quantization is projecting the original high-precision data to the discrete state space, of which deterministic and stochastic projections are the two main approaches used widely. The former projects the high-precision data to the nearest discrete state, while the latter projects the data to one of the two adjacent states with probability which is determined by the distance from the original data to the discrete states. Generally, deterministic projection is easier to handle, so that most references listed in Table \ref{Table-quantization} select it.

\subsubsection{Typical Practices}\quad

In the aspect of CNNs, we list a number of typical and latest practices with their best performance on CIFAR10 or ImageNet in Table \ref{Table-quantization}. Note that compression ratio of quantization is not considered because it relates directly to the bit-width, thus the storage saving is limited. As can be clearly observed, all the recent practices quantize W, and most works quantize both W and A, a small number of works quantize G,E, or U. It is worth to mention that the practice which quantizes W, A, G and E \cite{Wu_2018_WAGE} inspires some practices on semantic segmentation tasks in terms of accelerating corresponding encoder-decoder CNNs \cite{Poudel_2018_WAGE_1,Poudel_2019_WAGE_2}. We emphasize quantized objects here rather than other aspects of quantization, \emph{e.g.}, state distribution and state projection, is due to that the choice on which objects are quantized or not can influence the training and inference processes significantly. Particularly, quantized G and U can simplify the training very much because the data flow of back propagation can be all handled in integers. Moreover, Banner et al. \cite{banner2018scalable} propose the range Batch Normalization (BN) to greatly reduce the numerical instability and arithmetic overflow caused by the popular standard deviation-based BN, hence the BN can also be quantized. We optimistically believe that some entirely quantized DNNs will come soon and then mature and efficient integer neural networks should become the mainstream especially for embedded surroundings.

The amount of practices of quantized RNNs for visual tasks is few to the best of our knowledge, in comparison with natural language processing tasks \cite{courbariaux2015binaryconnect,rastegari2016xnor,hubara2018quantized,zhou2016dorefa,zhou2017balanced}. Unlike CNNs, RNNs are dynamic systems that reuse the weight and accumulate the activation error in temporal dimension, and furthermore, there are two dimensions of back propagation in RNNs, which are spatial (layer-by-layer) and temporal (step-by-step). These situations make it harder to clarify the training dataflow and quantization sensitivity. In fact, many quantization methods can be shared by both CNNs and RNNs \cite{courbariaux2016binarized,zhou2017balanced}, however, more applications of quantized RNNs for visual tasks should be established in consideration of the complexity of RNNs discussed above.

\subsection{Pruning}

Pruning can reduce the amount of weights or neurons, thus the memory and calculation costs are retrenched. However, the additional indices for indicating the location of non-zero elements, and the irregular access or execution pattern, become the two major drawbacks.

\subsubsection{Basic Method}\quad

\noindent \textbf{Problem Formulation}.
Similar to quantization, there are also two formulations which can describe pruning. The first one is direct and naive, \emph{i.e.}, utilizing some search algorithms for trained DNNs to find those ``unimportant" weights or neurons to prune, like
\begin{equation}
S(\bm{X}) = sparse(\bm{X})
\label{Sparse_search}
\end{equation}
where \(\bm{X}\) is the weight matrix  and \(S(\bm{X})\) is the new weight matrix after pruning. The function \(sparse(\cdot)\) is the search algorithm to pre-select the unimportant weights and neurons to be pruned. Such search approaches could be low-precision estimation \cite{lin2017predictivenet,song2018prediction}, negative activation prediction \cite{aklaghi2018snapea}, etc. Besides, hashing trick \cite{Weinberger2009Hashing} may also help to make weights in DNNs sparse \cite{Chen2015Hashing} even it seems to be weights sharing rather than a pruning approach in concept. Further, some other hashing methods may reveal similar abstract thought to normal pruning, \emph{e.g.}, using locality sensitive hashing (LSH) to collect neural nodes in active set and other neural nodes which are not in this set will not be computed during forward and backward calculations \cite{spring2017scalable}.

However, it is obvious that search algorithm may consume vast computing time for DNNs with large scale. Further, as the same reason for quantization, regarding the problem of pruning as optimization may be a more adaptive formulation to large DNNs because of the layer coupling. A general formulation of pruning in optimization can be described as \cite{wen2016learning}
\begin{equation}
\underset{W}{\text{min}} ~L_{0}(W) + \lambda\sum_{g=1}^G {\lVert \bm{W}^{(g)} \rVert}_{2}
\label{Sparse_optim}
\end{equation}
where \(W = \{\bm{W}^{(1)}, \bm{W}^{(2)}, \cdots, \bm{W}^{(G)}\}\) is the set of all weights in \(G\) different layers or parts; \(\lambda\) is a penalty parameter that affects the sparsity, and \(L_{0}(W)\) is the normal loss function of DNNs.

\begin{figure}
\centering
\includegraphics[width=0.42\textwidth]{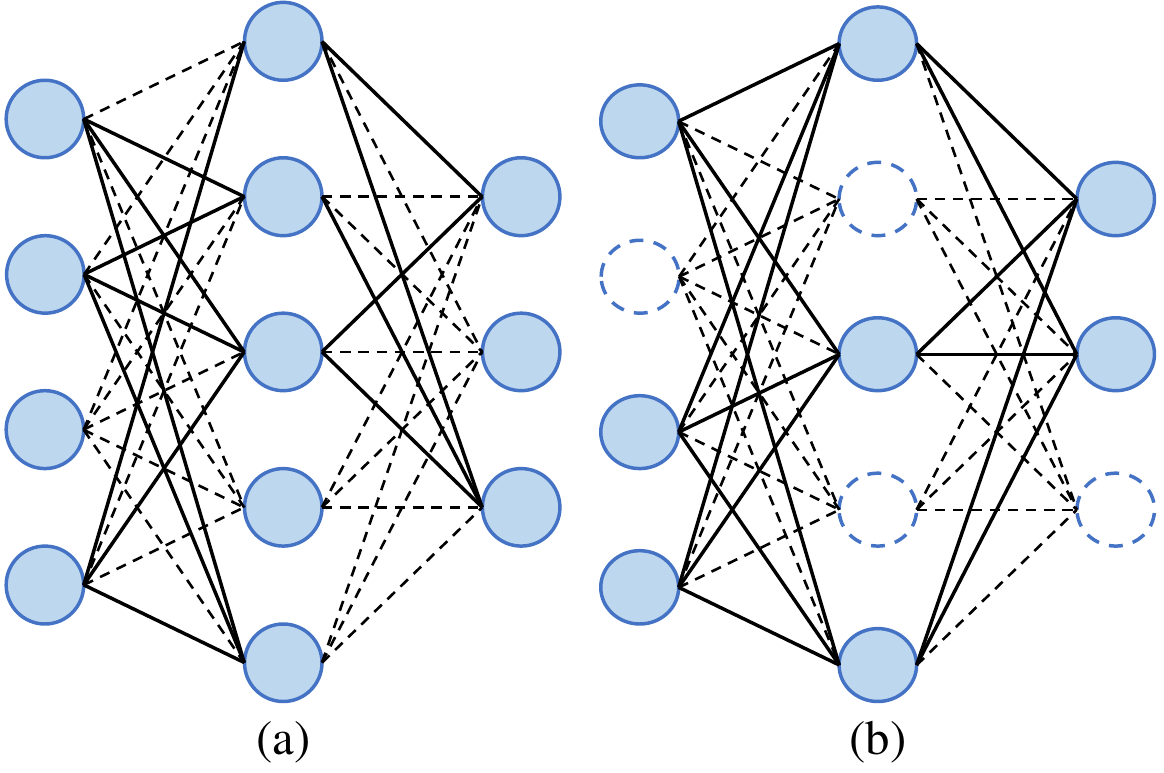}
\caption{Pruning objects: (a) weight pruning; (b) neuron pruning.}
\label{fig_sparse_method}
\end{figure}

\noindent \textbf{Pruning Objects}.
There are two typical pruning objects: weight pruning and neuron pruning, which are illustrated in Fig. \ref{fig_sparse_method}. The former one reduces the number of edges that can make weight matrices sparser, while the latter one reduces the number of nodes to  make weight matrices smaller. Evidently, the latter method may cause more accuracy loss than pruning weights only. Using ResNet50 as an example, according to Table \ref{Table-sparsifying}, weight pruning networks \cite{lin2017deep,chin2018layer,he2018amc,xu2018hybrid,lin2018synaptic} can exceed neuron pruning networks \cite{luo2017thinet,ye2018rethinking,luo2018autopruner} in terms of the performance of top-1 accuracy on ImageNet in most cases. Nevertheless, this gap has been reducing recently.

\noindent \textbf{Pruning Structure}.
Generally, the compute acceleration has a great deal to do with the sparse pattern, which is termed as pruning structure in this paper. It is well-known that the operation in one neural layer can be abstracted as matrix multiplication, thus the pruning structure can be described as the amount of zeros in matrix. Besides, the convolutional computation is commonly converted to the modality of GEneral Matrix Multiplication (GEMM) by lowering features and weight tensors to matrices \cite{chetlur2014cudnn}.

\begin{figure}
\centering
\includegraphics[width=0.45\textwidth]{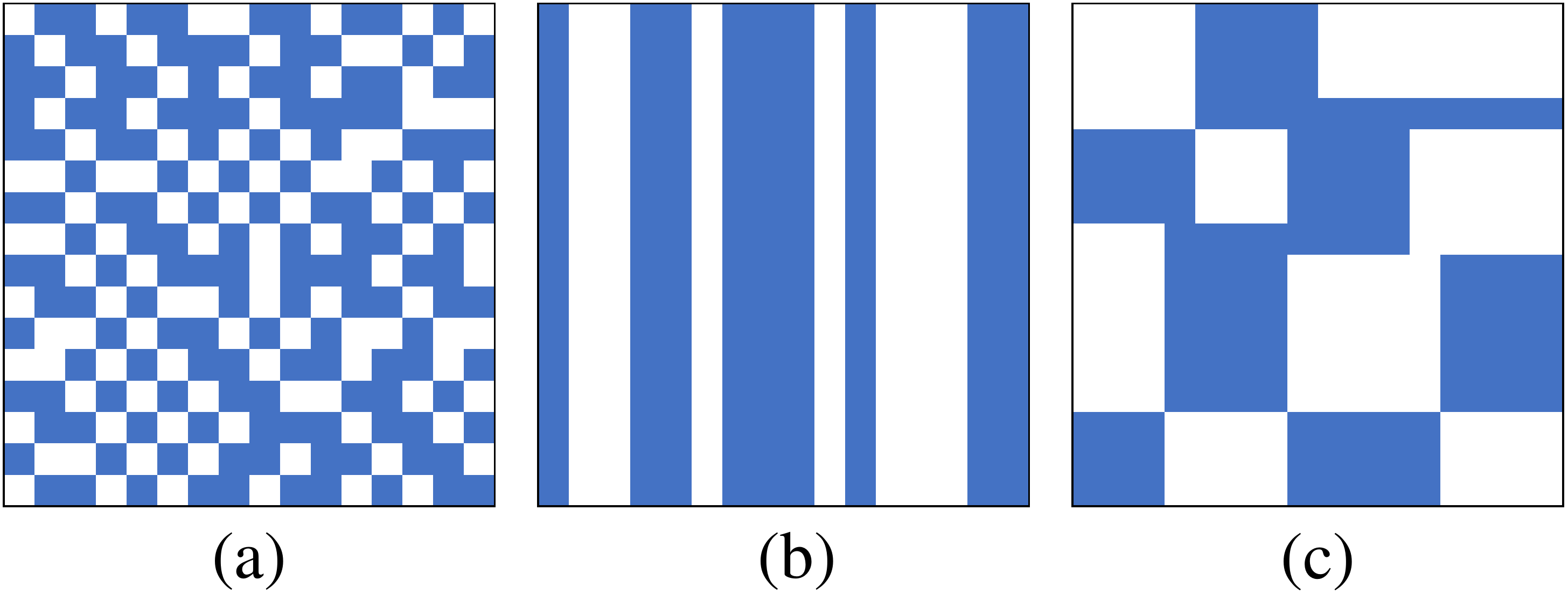}
\caption{Pruning structure: (a) element-wise; (b) vector-wise; (c) block-wise.}
\label{fig_sparse_structure}
\end{figure}

Fig. \ref{fig_sparse_structure} illustrates different pruning structures: element-wise, vector-wise, and block-wise. Note that different pruning grains produce different pruning structures. For instance, in weight pruning, kernel (discrete), fiber, or filter pruning produces vector sparsity, while channel or kernel (group) pruning produces block sparsity.

\subsubsection{Typical Practices}\quad

\begin{table*}
\caption{Typical practices of pruning for CNNs.}\vspace{3pt}
\label{Table-sparsifying}
  \centering
  \renewcommand\arraystretch{1.6}
  \resizebox{0.95\textwidth}{!}{
  \begin{tabular}{c | c c c c c c}
   \hline
   \hline
     References & Formulation & Object & Structure & Compression Ratio & Performance & Accuracy Loss \\
     \hline
     Prune or Not (2017) \cite{zhu2017prune} & Search & W & Element & \(8\times\) & ImageNet, InceptionV3, 74.6\% & 3.5\% \\
     Nest (2017) \cite{dai2017nest} & Search & W \& N & Element \& Block & \begin{tabular}{c} \(15.7\times\) \\ \(13.9\times\) \end{tabular} & \begin{tabular}{c} ImageNet, AlexNet, 57.24\% \\ ImageNet, ResNet50, 71.94\% \end{tabular} & \begin{tabular}{c} 0.02\% \\ 0.35\% \end{tabular} \\
     FDNP (2018) \cite{liu2018frequency} & Search & W & Element & \(22.6\times\) & ImageNet, AlexNet, 56.82\% & -0.24\% \\
     meProp (2017) \cite{sun2017meprop} & Search & N & Element & - & Mnist, MLP500\(\times\)2, 98.32\% & -0.23\% \\
     DGC (2017) \cite{lin2017deep} & Search & W & Element & \begin{tabular}{c} \(597\times\) \\ \(277\times\) \end{tabular} & \begin{tabular}{c} ImageNet, AlexNet, 58.2\% \\ ImageNet, ResNet50, 76.15\% \end{tabular} & \begin{tabular}{c} -0.01\% \\ -0.06\% \end{tabular} \\
     NISP (2017) \cite{Yu_2018_CVPR} & Optimization & N & Block & \begin{tabular}{c} \(1.51\times\) \\ \(1.78\times\) \end{tabular} & \begin{tabular}{c} ImageNet, GoogLeNet, - \\ ImageNet, ResNet50, - \end{tabular} & \begin{tabular}{c} 0.21\% \\ 0.89\% \end{tabular} \\
     ThiNet (2017) \cite{luo2017thinet} & Optimization & N & Block & \begin{tabular}{c} \(16.64\times\) \\ \(2.06\times\) \end{tabular} & \begin{tabular}{c} ImageNet, VGG16, 67.34\% \\ ImageNet, ResNet50, 71.01\% \end{tabular} & \begin{tabular}{c} 1\% \\ 1.87\% \end{tabular} \\
     Channel Prune (2017) \cite{he2017channel} & Optimization & N & Block & - & \begin{tabular}{c} ImageNet, ResNet50, 90.8\%(top-5) \\ ImageNet, Xception, 91.8\%(top-5) \end{tabular} & \begin{tabular}{c} 0.4\% \\ 1\% \end{tabular} \\
     Slimming (2017) \cite{liu2017learning} & Optimization & N & Block & \begin{tabular}{c} \(2.87\times\) \\ \(1.54\times\) \end{tabular} & \begin{tabular}{c} CIFAR10, DenseNet40, 94.35\% \\ CIFAR10, ResNet164, 94.73\% \end{tabular} & \begin{tabular}{c} -0.46\% \\ -0.15\% \end{tabular} \\
     ISTA (2018) \cite{ye2018rethinking} & Optimization & N & Block & \(1.89\times\) & ImageNet, ResNet101, 75.27\% & 1.13\% \\
     AutoPrunner (2018) \cite{luo2018autopruner} & Optimization & N & Block & \(3.33\times\) & ImageNet, ResNet50, 73.05\% & 3.1\% \\
     2PFPCE (2018) \cite{min20182pfpce} & Optimization & N & Block & \(8.33\times\) & CIFAR10, VGG16, 91.0\% & 1.98\% \\
     SFP (2018) \cite{he2018soft} & Search & W & Vector & \(1.43\times\) & ImageNet, ResNet101, 77.51\% & -0.14\% \\
     LCP (2018) \cite{chin2018layer} & Search & W & Vector & - & ImageNet, ResNet50, 75.28\% & 0.85\% \\
     AMC (2018) \cite{he2018amc} & Search & W \& N & Element \& Block & \(5\times\) & ImageNet, ResNet50, 76.11\% & 0.02\% \\
     NestedNet (2018) \cite{kim2018nestednet} & Search & W \& N & Element \& Block & \(2.74\times\) & CIFAR100, WRN-14-4, 74.3\% & 1.2\% \\
     Hybrid Prune (2018) \cite{xu2018hybrid} & Search & W & Element \& Vector & \(3.69\times\) & ImageNet, ResNet50, 74.32\% & 1.69\% \\
     Channel Prune (2018) \cite{hu2018novel} & Optimization & W & Vector & \(1.05\times\) & ImageNet, VGG16, 70.24\% & -2.08\% \\
     Crossbar-aware (2018) \cite{liang2018crossbar} & Optimization & N & Block & \(1.42\times\) & ImageNet, ResNet18, - & 1.65\% \\
     TETRIS (2018) \cite{ji2018tetris} & Optimization & N & Block & \(12.5\times\) & ImageNet, VGG16, 72.5\% & \(<\)0.1\% \\
     Joint Sparsity (2018) \cite{choi2018compression} & Optimization & W & Vector & \(2.8\times\) & ImageNet, ResNet18, 67.8\% & 0.4\% \\
     Synaptic (2018) \cite{lin2018synaptic} & Optimization & W & Vector & \(4.34\times\) & ImageNet, ResNet50, 74.68\% & 0.62\% \\
     ADMM (2018) \cite{zhang2018adam} & Optimization & W & Any Structure & \(16.1\times\) &  ImageNet, AlexNet, 57.42\% & 0.2\% \\
     Importance (2019) \cite{Molchanov_2019_Importance} & Search & W \& N & Element \& Block & \(3.29\times\) & ImageNet, ResNet101, 74.16\% & 3.21\% \\
     SSR (2020) \cite{Lin_2020_SSR} & Optimization & N & Block & \(2.13\times\) & ImageNet, ResNet50, 71.47\% & 3.65\% \\
     Rewinding (2020) \cite{Renda_2020_Rewind} & Search & W \& N & Element \& Block & \(5.96\times\) & ImageNet, ResNet50, 76.17\% & $\sim$0\% \\
   \hline
   \hline
  \end{tabular}}
  \vspace{-5pt}
\end{table*}

Here we let W and N denote weight pruning and neuron pruning respectively. In the aspect of CNNs, according to Table \ref{Table-sparsifying}, the works that prune W and those prune N are close in terms of quantity, and just a few practices have considered both of them (W and N) \cite{dai2017nest}. We emphasize pruning objects here is due to that, as mentioned before, pruning only weight or neuron may influence the performance more obviously than other aspects, \emph{i.e.}, formulation and structure. From Table \ref{Table-sparsifying}, in terms of performance, it cannot be deemed absolutely that optimization is better than searching for large DNNs until now. One certain advantage of optimization is efficient training and avoiding some time-consuming searching procedures. For pruning, element structure can be helpful for achieving higher compression ratio, especially the case of DGC \cite{lin2017deep}. Relatively, other structures may bring significant acceleration of computing, particularly filter (vector) and channel (block) pruning in convolutional kernels. On the whole, pruning on convolutional kernels is much more difficult than that on fully connected layers.

In the aspect of RNNs, there are two typical image captioning tasks, among which one prunes W \cite{dai2018grow} and the other prunes N \cite{zhu2018structurally}. The former has achieved \(13.12\times\) compression ratio on MS COCO dataset with 2.3 improvement of CIDEr score. The latter has also used MS COCO dataset and got only 0.4 reduction of BLUE-4 score with 50\% sparsity. Comparatively speaking, \wang{it might imply} that there are still lots of potentials for visual RNNs with pruning to get further advanced.

\subsection{Discussions and the Derived Joint Compression}\label{subsec:joint_compress}

In fact, each compression method has its own characteristics which other compression methods do not have. Before listing recent practices of joint compression, we present here our observations and thoughts about what features make each compression method worthy and unique based on the foregoing content in this section.

\subsubsection{Main Characteristic of Each Method}\quad

\noindent \textbf{Compact Networks}.
As discussed before, there are two levels of compact designing, one is delicate cell and the other is NAS algorithm. According to Table \ref{tab-nas-imagenet}, NAS shows superior performance compared with human designed networks in both accuracy and storage saving. The most critical point is that, executing time of NAS algorithm has been shortened a lot nowadays \cite{liu2018darts}, such situation makes NAS \emph{promising and generic} for embedded applications with various or changing surroundings though there are still a lot of detailed works to do towards real applications. While other compression methods, \emph{i.e.}, decomposition, quantization and pruning, are still lack of enough flexibility, because a single specific network architecture is hard to be applied to all kinds of datasets.

\noindent \textbf{Tensor Decomposition}.
In early studies of network compression, tensor decomposition was the only approach that supports the so called \emph{in situ} training \cite{Alibart_2013_ExInSitu}, which means training a new model from scratch. Meanwhile, quantization and pruning needed pre-training in most cases to discover the distribution of weights to quantize or prune further. However, new studies of quantization \cite{jung2018joint} and pruning \cite{liu_2019rethinking_ICLR} have made up this short slab. Moreover, both of these two reports conclude that \emph{in situ} training could perform better than fine-tuned models. Even so, tensor decomposition still appears to be the most powerful in the aspect of compression ratio particularly for RNNs according to Table \ref{Table-TT-Application}. Currently, the most unique characteristic of tensor decomposition is that various tensor network formats may have some inner linkages to DNN architectures, thus \emph{theoretical explanation} of efficient DNNs may be explored in this kind of compression. Chien et al. \cite{ChienTuckerLayer} regard the whole Tucker decomposition process in Equation (\ref{ClassicalTensorDecomposition}) as a neural network connection. Cohen et al. \cite{CohenHT,CohenHT2} use HT decomposition to explain the depth efficiency of DNNs. Chen et al. \cite{ChenTuckerBTD} find BTD can describe various bottleneck architectures in ResNet and ResNeXt. Li et al. \cite{Li_2018_MERA} propose a new DNN architecture based on MERA tensor network, which in actual is a kind of renormalization group (RG) transformation \cite{Evenbly_2009_MERARG}. In this sense, \wang{it is expected that} the studies of compressing DNNs with tensor decomposition can build a bridge between the experiments and the system of theories about DNNs.

\noindent \textbf{Data Quantization}.
Although data quantization is not very good at reducing model complexity (in terms of the number of parameters) of DNNs compared with decomposition and pruning, it still has a significant advantage of \emph{computing acceleration and friendly deployment of embedded hardware}, which are guaranteed by the low bit data flow in quantized DNNs. The authors implemented WAGE \cite{Wu_2018_WAGE} on FPGA platform, further found 8 bit models perform \(3\times\) faster in speed, \(10\times\) lower in power consuming, and \(9 \times\) smaller in circuit area than 32 float point models \cite{Yang_2019_WAGEU}. Although pruning with appropriate structure may also help to reduce computation complexity, bit operations in quantized DNN are still more adaptive for hardware environments.

\noindent \textbf{Pruning}.
According to Table \ref{Table-TT-Application} and Table \ref{Table-sparsifying}, pruning has a better capability of \emph{accuracy maintenance or even improvement}, though seemingly other methods like tensor decomposition has more potential power to gain higher compression ratio. However, accuracy loss of compressed DNNs under decomposition is hard to avoid especially for CNNs at present. For example, the performance of VGG16 on CIFAR10 with pruning can achieve amost 0.15\% accuracy lift in the new study \cite{liu_2019rethinking_ICLR}, while the accuracy loss is hard to make up even the ranks are set very high based on TT convolutions \cite{Wang_2020_TT3DCNN}. Additionally, two main advantages of decomposition, \emph{i.e.}, higher compression ratio and \emph{in situ} training, can be separately obtained by \cite{lin2017deep} and \cite{liu_2019rethinking_ICLR} in the aspect of pruning. However, similar to NAS, data structure of pruned DNNs may present complex and chaos, which is unfriendly to embedded applications and corresponding theory explanation.

\subsubsection{Joint Compression}\quad

Considering different characteristics of each compression approach, recently some researchers have made their applications involve more than one class of method besides using individual compression method, termed as joint compression in this paper. Corresponding visual recognition works are illustrated in Table \ref{tab:Joint-compression}. The joint compression has great potential for higher compression ratio. For instance, compression ratio of 89\(\times\) of AlexNet on ImageNet (58.69\% accuracy) \cite{zhou2017incremental}, 28.7\(\times\) of ResNet18 on ImageNet (66.6\% accuracy) \cite{choi2018compression}, and 1910\(\times\) of LeNet5 on Mnist (98\% accuracy) \cite{ye2018progressive}. 

However, researchers should pay more attention to maintaining the model accuracy in joint compression through sufficient analyses and comprehensions about respective characteristic of each compression component. For example, since there might be a projection between the structure of tensor decomposition and the architecture of DNN as discussed above, \wang{the decomposition method appears to be perpendicular to the other methods to some extent}, so combining it with quantization or pruning is a promising direction which still needs further researching according to Table \ref{tab:Joint-compression}.

\begin{table}
\caption{Extant works of joint compression for visual recognition tasks.}\vspace{3pt}
\label{tab:Joint-compression}
  \centering
  \renewcommand\arraystretch{1.6}
  \resizebox{0.49\textwidth}{!}{
  \begin{tabular}{c | c c c c }
   \hline
   \hline
    References & Compact & Decompose & Quantize & \wang{Prune} \\
    \hline
  Deep Compression (2015) \cite{han2015deep} & - & - & \checkmark & \checkmark \\
   \hline
  SCNN (2015) \cite{liu2015sparse} & - & \checkmark & - & \checkmark \\
   \hline
   Force Regularization (2017) \cite{wen2017coordinating} & - & \checkmark &  & \checkmark \\
   \hline
   INQ (2017) \cite{zhou2017incremental} & - & - & \checkmark & \checkmark \\
   \hline
   Quantized Distillation (2018) \cite{polino2018model} & \checkmark & - & \checkmark & - \\
   \hline
   VNQ (2018) \cite{achterhold2018variational} & - & - & \checkmark & \checkmark \\
   \hline
   Joint Sparsity (2018) \cite{choi2018compression} & - & - & \checkmark & \checkmark \\
   \hline
   Regularization (2018) \cite{choi2018learning} & - & - & \checkmark & \checkmark \\
   \hline
   ADMM (2018) \cite{ye2018progressive} & - & - & \checkmark & \checkmark \\
   \hline
   PQASGD (2018) \cite{jiang2018linear} & - & - & \checkmark & \checkmark \\
   \hline
   DNNC (2019) \cite{Tzelepis_2019_DNNC} & - & - & \checkmark & \checkmark \\
   \hline
   \hline
  \end{tabular}}
\end{table}

\section{On Recognition: Inference and Generalization}
\label{sec:generalization_inference}

Efficient training can allow learning from more data, with more parameter tuning or more complete architecture search, which can all lead to a better trained model. However, it is also critical to make the model run efficiently on affordable or existing devices and easily transfer to new data/tasks. In many cases, training may be done just once or only at one place (\emph{e.g.} on the cloud) with powerful computational resources, but run-time inference has to be done at cost-sensitive and thus resource-limited edge computing devices. Therefore, in some sense, efficient run-time inference is a more serious issue when real applications are concerned. In this section, we focus on the fast run-time inference for model deployment at testing stage and transfer learning efforts for making the models easily generalized to new scenarios/tasks without costly retraining.

\subsection{Fast Run-time Inference}
\label{subsec:network_centric_compression}

Efficiency is not only a big concern for training deep learning models, but also rather important and sometime critical for the deployment at users' end in real applications. Therefore, how to accelerate run-time inference with limited resources is an indispensable issue of great importance for industrial applications. This is even more critical for applications that requires real-time or even faster responses, such as automatic recognition for autonomous driving.

There have already been a rich literature on accelerating run-time inference with DNNs. We roughly categorize them into two groups based on the difference of focus: \emph{data-aware acceleration} and \emph{network-centric compression}, and detail their recent advances and trends as follows.

\subsubsection{Data-aware acceleration}\quad

During run-time inference, in many cases there is no need to check all the input data for generating the final recognition results, especially for the data which may include much irrelevant or redundant information, such as videos. Therefore, efforts on reducing or avoiding the computation on the irrelevant/redundant parts of the data are very important for fast inference.

A simple yet very helpful direction is exploring the similarity of the intermediate feature maps of two consecutive video frames for reducing redundant computation. A representative earlier work is the one called deep feature flow \cite{Zhu_2017_CVPR}. It runs full expensive convolutions only on sparse key frames and then propagates their deep feature maps to other frames via a flow field. Significant speedup was achieved as flow computation is relatively faster than full convolution. This work got extended to a more unified framework \cite{zhu2018towards} which is proved to be faster, more accurate and more flexible. The framework contains three main components: sparsely recursive feature aggregation for ensuring both efficiency and feature quality, spatially-adaptive partial feature updating for improving the quality of features from non-key frames, and temporally-adaptive key frame scheduling for more efficient and better quality key frame usage. However, these two works are still computationally expense, as they rely on per-pixel flow computation, which is a heavy task.  

Recently, Pan et. al. \cite{pan_2018_CVPR} \ywunew{have explored another direction.} They proposed a novel recurrent residual module (RRM) which only do dense convolution on the first frame and have the following frames fed into a sparse convolution module which only extract information from the difference images of neighboring frames. The sparse convolution has no bias term and it shares the same filter banks and weights with dense convolution. After enhancing the sparsity of the data (by difference images), a general and powerful inference model which is called EIE (Efficient Inference Engine) \cite{han_2016eie_ISCA} is adopted to do the inference efficiently according to the dynamic sparsity of the input. In usage, the video is split into several chunks which can get processed with RRM-equipped CNN in parallel. Good results (speedup) have been observed on object detection and pose estimation in videos, and the model shall also be applicable to other visual recognition task for videos. Since it only explores the natural sparsity of data, it can ensure no accuracy loss during the speedup.

\begin{table*}
\begin{center}
\caption{Recent representative network compression works for fast run-time inference. Please refer to the text for details about the Key Property Indicators.}
\begin{adjustwidth}{-0.5cm}{-0.5cm}
\begin{scriptsize}
\begin{tabulary}{\linewidth}{|p{.08\textwidth}|p{.18\textwidth}|p{.14\textwidth}|p{.05\textwidth}|p{.05\textwidth}|p{.05\textwidth}|p{.05\textwidth}|p{.05\textwidth}|p{.18\textwidth}|}
\hline
\multirow{2}{*}{\small \textbf{Reference}} & \multirow{2}{*}{\small \textbf{Key Idea}} & \multirow{2}{*}{\small \textbf{Main Approach}} & \multicolumn{5}{c|}{\small \textbf{Key Property Indicator}} & \multirow{2}{*}{\small \textbf{Pros\&Cons (if known)}} \\
\cline{4-8}
&  &  & General or Specific? & Static or
 Dynamic? & Easy User Control? & Input 
Adaptive? & Resource
 Aware? & \\
\hline

Liu et. al. 2017 \cite{liu2017learning} & directly enforcing channel-level sparsity & scaling factors and sparsity-induced penalty & General
& Static & Yes & No & Yes & \textbf{Pros.}: Simple and general.
\textbf{Cons.}: Static (compression tied up training). \\
\hline
Lin et. al. 2017 \cite{lin_2017_NIPS};
Rao et. al. 2018 \cite{rao_2018_runtime_TPAMI} & model the
pruning of each convolutional layer as a Markov decision process (MDP) & reinforcement learning & Seem to be general &  Dynamic & Yes & Yes & Yes & \textbf{Pros.}: Good properties.
\textbf{Cons.} Overhead on pruning may be high. \\
\hline
Mullapudi et. al. 2018 \cite{mullapudi_2018_CVPR} & encourage components to be specialized and perform component selection during inference & a new network with multiple specialized branches, a gate for branching, and a combiner for aggregating. & Seem to be general &  Dynamic & Yes & Yes & Yes & \textbf{Pros.}: Intuitive, good properties.
\textbf{Cons.} Overhead on training may be high. \\
\hline
Gao et. al. 2019 \cite{gao_2019_ICLR} & use a low-overhead extra component to predict convolutional channels' saliency for prunning & a new piecewise differentiable and continuous
function for saliency prediction & Seem to be general &  Dynamic & No & Yes & No & \textbf{Pros.}: Simple, fast, currently lowest accuracy loss.
\textbf{Cons.} not directly resource-aware.  \\
\hline
Yu et. al. 2019 \cite{yu_2019slimmable_ICLR} & train a shared network with different widths & switchable batch normalization & General
 (verified) &  Dynamic & Yes & No & Yes & \textbf{Pros.}: Simple, clean, well-motivated, fast and easy to use.
\textbf{Cons.} Not input-adaptive. \\
\hline
Chang et. al. 2019 \cite{chang_2019urnet_arXiv} & using a Conditional
Gating Module (CGM) to determine the use of each
residual block according to the input image and the desired
scale & Conditional
Gating Module (CGM) & Specific to residual networks &  Dynamic & Yes & No & Yes & \textbf{Pros.}: Simple, clean, fast, and easy to use.
\textbf{Cons.} Not input-adaptive, and specific to residual networks\\
\hline
Zhang et. al. 2019 \cite{zhang_2019recurrent_arXiv} & cost-adjustable inference by varying the unrolling steps of Recurrent convolution (RC), with independently learned BN layers & Recurrent convolution (a particular kind of RNN) & Specific to RC networks &  Dynamic & Yes & No & Yes & \textbf{Pros.}: New approach, interesting idea.
\textbf{Cons.} Not input-adaptive, and specific to  RC networks\\
\hline
Liu et. al. 2018 \cite{liu_2018_MobiSys} & select a combination of compression techniques
for an optimal balance between
user-specified performance goals and resource constraints & reinforcement learning & General
 (verified) &  Dynamic & Yes & No & Yes & \textbf{Pros.}: systematic, application oriented.
\textbf{Cons.} Not input-adaptive, and the overhead can be high\\
\hline
Fang et. al. 2018 \cite{fang_2018nestdnn_MobiCom} & enables each DNN (by making it a multi-capacity model) to offer flexible resource-accuracy trade-offs, and then do resource-aware scheduling & a greedy heuristic approximation for optimizing MinTotalCost and MinMaxCost scheduling schemes & General
 &  Dynamic & Yes & No & Yes & \textbf{Pros.}: highly application oriented, general, little overhead.
\textbf{Cons.} Not input-adaptive; the overall solution looks complicated.\\
\hline
Pan et. al. 2018 \cite{pan_2018_CVPR} & using the similarity of the intermediate feature maps of two consecutive frames to largely reduce the redundant computation & the proposed novel Recurrent Residual Module & Video-specific
 &  Dynamic & No & Yes & No & \textbf{Pros.}: properly explored redundancy in data, no accuracy loss.
\textbf{Cons.} not user-controllable and not resource-aware.\\

\hline
\end{tabulary}
\label{table:inference_compression}
\end{scriptsize}
\end{adjustwidth}
\end{center}
\end{table*}

Besides information redundancy in consecutive frames, irrelevant information may largely exist for object detection in videos, as objects often occupy only a small fraction of each video frame. Therefore, an intuitive acceleration strategy is to do dense full processing on only a few frames and make use of the spatio-temporal correlation among nearby frames to save computation on the other frames. A representative work reallocates the computation over a scale-time space called Scale-Time Lattice \cite{Chen_2018_CVPR}. It performs expensive detection sparsely on key frames and propagates the results across both scales and time with substantially cheaper networks, by making use of the strong corrections among object scales and time. Together with some other minor novel components (\emph{e.g.} a network for temporal propagation and an adaptive scheme for keyframe selection), the work achieved better speed-accuracy tradeoff than previous works. Another representative work called Spatiotemporal Sampling Network (STSN) focuses on feature level propagation across adjacent frames. STSN is mainly about deformable convolutions across time, which is optimized directly with respect to video object detection performance. The approach has a natural robustness to occlusion and motion blur, which are two key challenges in detecting objects in videos.

\subsubsection{Network-centric compression}\quad

The main contents in the last section focus more on the theoretical methodology and training, meanwhile, for the practical scenarios, a lot of efforts have been made on general network-centric compression for fast inference. A comprehensive survey may have to cover several dozens of publications. Due to the scope and page limits of this paper, only a few representative ones for showing the recent advances and trends are covered here, as briefly introduced and compared in Table \ref{table:inference_compression}. 

Besides pointing out their key ideas, main approaches, advantages and disadvantages, we provide a group of Key Property Indicator (KPI) for briefly evaluating them in the perspective of important factors for real applications. There are totally five indicators, with the following detailed explanations.
\begin{enumerate}[label=\Roman*]
    \item ``\textbf{General or Specific?}'': whether the model is a general one that can be applied to the compression of any networks, or it is specific to some data types, tasks, or network structures.
    \item ``\textbf{Static or Dynamic?}'': ``static'' means that the compression has to be done together with the training of original models and any changes to the compressed model has to be accompanied with a retraining of the original model, so that once a compression model is optimized, it is static; ``dynamic'' means that the compression model can be changed dynamically on demand without requiring a retraining of the original model. In the last two years, there is a clear trend towards shifting to dynamic compression, so most of the references in Table \ref{table:inference_compression} are about dynamic models and only one representative of static model is listed.
    \item ``\textbf{Easy User Control?}'': whether the compression can be easily controlled by the users or not, such as using one or very few hyper-parameters for an intuitive controlling of the compression/speedup rate.
    \item ``\textbf{Input Adaptive?}'': whether the compression is made adaptive to the input data or not. When it is input adaptive, the network can be changed for each specific input so that the inference time may be greatly reduced for easier inputs.
    \item ``\textbf{Resource-Aware?}'': whether the compression is aware of the available resources (including resources for storage, transmission and computation) or not. Fewer and weaker resources shall led to higher compression rate, and vice versa. Resource types shall also matter in detailed control of the compression.
\end{enumerate}

There are a few findings in our study, which are worth mentioning.
\begin{itemize}
    \item Like the channel-wise sparsity work proposed by Liu et. al. \cite{liu2017learning} or the slimmable neural networks proposed by Yu et. al. \cite{yu_2019slimmable_ICLR}, \textbf{there is a clear trend that the network pruning focuses more and more on whole channels or blocks} \cite{chang_2019urnet_arXiv} as such structural pruning is GPU-friendly (can be exploited by GPUs) and allows the acceleration to work on dense operations of fewer components, instead of many sparse individual weights.
    \item \textbf{Models which can be easily controlled by users (``Easy User Control'') are usually also ``Resource-Aware''}, as there is a common assumption that user can control the compression according to the actual situation of resources. However, in real cases, the state of resources can be dynamic even for a specific device instance (\emph{e.g.} someone's smartphone), as all the factors of storage, transmission and computation can be changing over time. There can be multiple processes/applications running at the same time and the internet connection speed is hard to be consistent. Therefore, \textbf{instead of asking user to specify the compression rate, having a model to dynamically decide it is a more practical and also more promising choice, which is widely ignored and far from being explored}. Fang et. al. \cite{fang_2018nestdnn_MobiCom} present an inspiring work in this direction.
    \item \textbf{Being ``Resource-Aware'' and being ``Input-Adaptive'' seem hard to be achieved at the same time}, as the former is about overall compression which can be controlled by users, while the latter is about automatically adjusting of compression based on each individual input data instance. However, these two can be solved from different aspects and there is no conflict between them. Therefore, \textbf{integrating both factors (``Resource-Aware'' and ``Input-Adaptive'') is a promising direction worth more investigation}, and there are also a few good examples \cite{lin_2017_NIPS,rao_2018_runtime_TPAMI,mullapudi_2018_CVPR}. 
\end{itemize}

\subsection{Generalization via Transfer Learning}
\label{subsec:generalization_TL}


It is well-known that transfer learning has been widely utilized for DNN-based visual recognition. Leveraging weights of a good model pre-trained on a large well-annotated dataset like ImageNet~\cite{imagenet_cvpr09} has been a standard way for a great range of visual tasks, motivated by the encouraging results of even the most naive way of transferring ~\cite{ImageRep_CNN_Features_Baseline_CVPRW14}. Transferring the learned knowledge of the source task to a related task could significantly reduce the training time while at the same improving the recognition performance, especially for few-shot recognition problems.

Transfer learning has been playing a critical role in reducing annotation burdens for DNNs-based applications. Since deep learning's performance heavily depends on the quantity and quality of data, in a deep learning project pipeline, a huge amount of time could be spent on data collection and annotation. Therefore, a neural network model is more efficient when it can use less manually annotated data to reach good results (\emph{e.g.} comparable with those consume much larger amount of labeled data).

To reduce manual annotation efforts on the target task, usually either synthetic data or data from similar tasks are employed. Compared with the target data domain, however, the available source data may have different feature domains. Ignoring the domain gap and directly applying the model trained on the source data or performing co-training on both data domains could compromise the neural network's performance. Therefore, domain adaptation, an important and widely-studied branch of transfer learning, has been utilized to bridge the gap between different data domains~\cite{pan2009survey}. The successful applications cover many visual tasks including image classification~\cite{qiao2018deep}, pose estimation~\cite{rad2018feature}, image segmentation~\cite{watanabe2018multichannel}, object detection~\cite{saito2018strong}, and image captioning~\cite{chen2017show}, etc. 

Recently, there have been increasing interests in deeper understandings of transfer learning for DNNs. A theoretical study on the transferability of deep representations is given in~\cite{Transfer_liu2019understanding}, and a comprehensive empirical study is done in~\cite{Transfer_Kornblith_2019_CVPR} with many valuable findings. It is great to see that better pre-trained models generally perform better and transferring with fine-tuning is better than training-from-scratch for most of the studied diverse image recognition tasks with many smaller-scale datasets. There is an important message on the efficiency issue related to transfer learning: for achieving the same accuracy (``90\% of the maximum odds of correct classification achieved at any number of steps''), fine-tuning a pre-trained model is averagely 17 times faster than training from scratch (the reason for that is discussed in ~\cite{Transfer_liu2019understanding}.). Since the dataset (\emph{e.g.} ImageNet) is usually more than 20 times larger than the target dataset, if the pre-training time is also taken into account, the total time for training from scratch is likely to be shorter. However, when both the DNN model and the dataset for pre-training is public (\emph{e.g.} ImageNet), which is often the case, directly borrowing a pre-trained model is a better choice for efficiency, as the time for pre-training is already a sunk cost paid by some others. Reusing it has no time cost and it is also a good choice as these models are usually carefully optimized when they were published online.

For an even faster transferring (with possibly a better performance) than fine-tuning, recently some pioneering studies have proposed to adapt the pre-trained model to a specific target task/domain together with network structure compression so that both the full model adaption and fewer parameter learning can be achieved at the same time. Learning fewer parameters during transferring is believed to be more efficient than naive full model fine-tuning.  In~\cite{Transfer_Rosenfeld_2018_PAMI}, Deep Adaptation Modules (DAM) are introduced to constrain newly learned filters to be linear combinations of existing ones and force significant parameter size reduction (typically 13\%, dependent on network architecture), and when coupled with standard network quantization techniques, they are said to be able to further reduce the parameter cost to around 3\% of the original full pre-trained model. A similar idea is proposed in~\cite{Transfer_Rebuffi_2018_CVPR}, based on the parametrization of universal parametric families of neural networks covering specialized problem-specific models, with just small differences among them. Besides directly changing the pre-trained model, there is also a new idea of fixing it during the transferring and adding extra adapter modules to it for adapting to different target tasks~\cite{Transfer_houlsby19a_2019_ICML}. During transferring, only the adapter module is learned and the module is made to be much lighter than the original pre-trained model so that the additional learning can be very efficient. In the paper \cite{Transfer_houlsby19a_2019_ICML}, extensive experiments show that with only 3.6\% parameters (of the original pre-trained model), the performance can go beyond 99.6\% of fine-tuning the whole pre-trained model. Though the last idea on appending has only been tested on NLP, the model looks general and therefore similar idea may work for visual recognition tasks too.

\section{Summary and Discussions}
\label{sec:summary_discussions}

\subsection{Summary of Recent Advances}

\begin{figure}
\centering
\includegraphics[width=1.0\columnwidth]{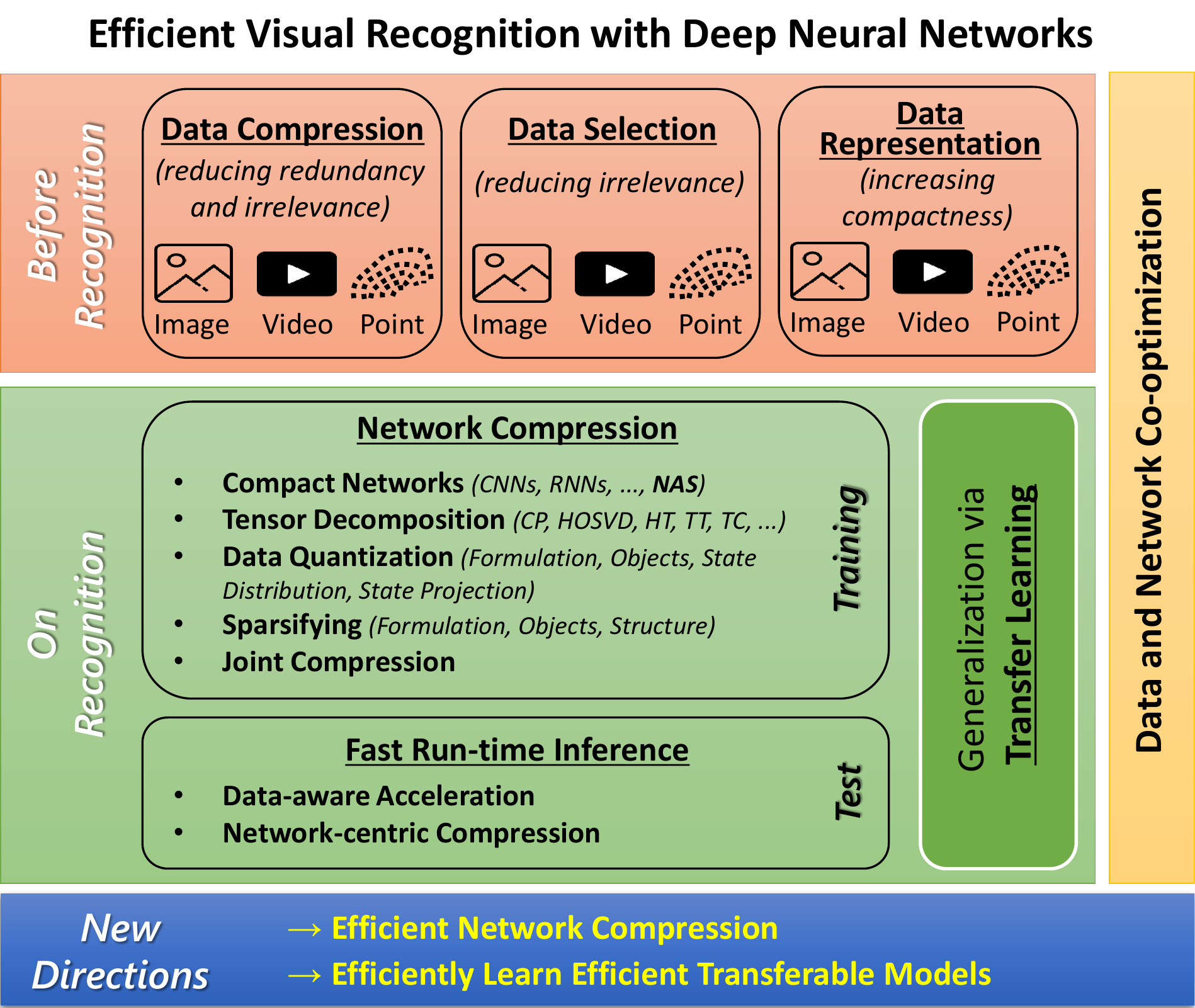}
\caption{Overview of the recent advances and new directions.}
\label{fig:summary}
\end{figure}

As detailed in the former sections and also summarized in Fig. \ref{fig:summary}, to design an efficient visual recognition model or system, one may put efforts on data processing (compression, selection and representation) which is usually data type specific, network compression for training, fast run-time inference and efficient transfer learning, as already explored by the rich literature in the past few years. 

\textbf{An important message is that a lot of things can be done on the data side or having it taken into account when people design new efficient models}, which is not only important for the recognition performance, but also critical for ensuring that the network compression fits the data well. 

\subsection{An Important New Frontier: Data and Network Co-optimization}

Recently, some methods actually have collaboratively optimized the data and networks together, although we reviewed them separately as `\textbf{Before Recognition}'' and ``\textbf{On Recognition}'' above. For example, in a previously cited paper \cite{torfason2018towards}, the network is cascaded by a data compression network and a classification network and trained in an end-to-end way, thus the compression network can compress images most efficiently, retain classification information, and also improve classification's efficiency. Similarly, in \cite{wu2018compressed}, the compressed video is used as the input of the action recognition network, and the decoding function is assigned to the recognition network through training, therefore the network learns the abilities for decoding and recognition at the same time through an end-to-end training.

In fact, training end-to-end has been a common desire for DNN-based solutions, however, even though joint compression and recognition is already made possible, so far the existing solutions have still treated them as two separate yet linked modules rather than a single fused model. Comparing with linking them for co-optimization, fusing them as a whole model may be more helpful for developing more computationally efficient DNN models with minimum redundant computations for real applications, even though that may lead to more space consumption due to the uncompressed data which is nowadays relatively easier to handle. Since we have not found any corresponding practices to the best of our knowledge, here we propose two possible directions for such a unified model design.

One direction is to explore quantization: linking data quantization and network quantization. As discussed earlier in subsection \ref{subsection:quantization} (esp. Fig. \ref{fig_q_object} and Table \ref{Table-quantization}) and subsection \ref{subsec:joint_compress}, quantization can be done for the whole data flow inside DNNs and it is quite superior for adapting to various hardware platform, thus it is closer to real applications. Meanwhile, currently most visual data are originally represented by low bit integers thanks to the advancement of digital visual sensors, which make data quantization convenient and straight-forward. The main barrier for linking these two is that currently the raw visual data format may not directly match to quantized DNN data types. Therefore, we think necessary and highly valuable future efforts should be either on transforming the representation of raw data towards that of the quantized DNN models or going to the extreme to make the new visual sensors being able to produce various or more easily adjustable raw data formats for the integration with DNN quantization.

\begin{figure}
\centering
\includegraphics[width=0.35\textwidth]{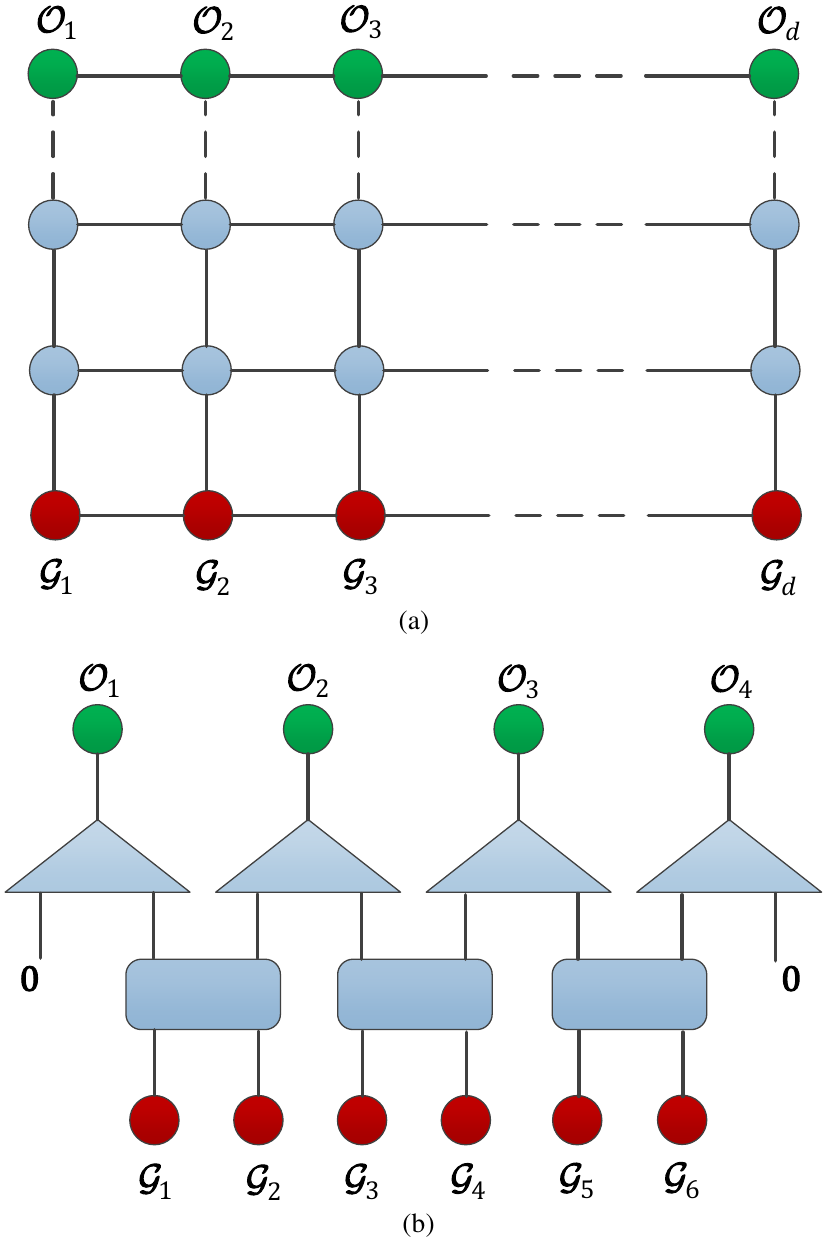}
\caption{Input data (denoted by the red nodes $\bm{\mathcal{G}}_1$, $\bm{\mathcal{G}}_2$, $\cdots$) and weights in DNNs may be uniformly modeled as a tensor network , with typical examples such as (a) PEPS \cite{Schuch_2010_PEPS} and (b) MERA \cite{Evenbly_2009_MERARG}.}
\label{fig_peps_mera}
\end{figure}

The other direction is to extend tensor network for covering the input data. Tensor networks can inherently describe linear transformations, \emph{e.g.}, matrix production and tensor contraction. Meanwhile, a DNN architecture is generally similar to a tensor network except all the nonlinear activation functions. Therefore, tensor network can at least inspire us to develop some new DNN architectures by analysing probable relationship between tensor networks and {DNNs}, even though the strict mapping from tensor network to DNN may be difficult. Thus, compressed input data and a compressed DNN architecture can be both put into a system which is similar to a tensor network. For example, if the input data and the weights in a DNN are both approximated in TT format, the whole system can be expressed like a projected entangled-pair states (PEPS) tensor network \cite{Schuch_2010_PEPS} as drawn in Fig. \ref{fig_peps_mera}(a) where red and green nodes illustrate input and output data respectively, and existing efficient computing algorithms, \emph{i.e.}, Algorithm 5 in \cite{OseledetsTTInvent2}, may help to accelerate this kind of DNNs. Further, if one layer of the DNN is designed like multi-scale entanglement renormalization ansatz (MERA) \cite{Evenbly_2009_MERARG}, higher ability of expression may be obtained as presented in Fig. \ref{fig_peps_mera}(b). It is easy to observe that DNN in the MERA architecture has stronger local correlation than PEPS, \emph{e.g.}, the content of \(\bm{\mathcal{O}}_2\) in Fig. \ref{fig_peps_mera}(b) comprises the information from \(\bm{\mathcal{G}}_1\), \(\bm{\mathcal{G}}_2\), \(\bm{\mathcal{G}}_3\) and \(\bm{\mathcal{G}}_4\). Through such efforts, the efficiency of data representation and network may be both optimized via the tensor network models with theoretical supports.

\subsection{Unexplored Yet Promising New Directions}

Though important and attractive recently-emerged directions have already been discussed in the former sections when individual topics are introduced, there might be still some unexplored yet promising directions, which may be worth mentioning. Within them, the following two are believed by us to be most valuable, though it is still challenging to work on.

\subsubsection{Efficient Network Compression}\quad

Network compression is now a vibrant research subject, especially orienting to visual recognition neural networks which are always large-scaled due to high-dimensional raw visual data. However, according to our investigation, we summarize several promising directions of compression methods below which may promote the miniaturization of DNNs further.

\begin{itemize}
    \item \textbf{NAS may become a promising or even requisite approach especially for embedded visual applications.} Such a particular approach is very suited to the application scenarios with variable surroundings, \emph{e.g.}, fault diagnosis based on visual information, ground object identification based on aerial photography, etc.
    
    \item Some tensor decomposition methods have not been studied adequately due to their complexity, \emph{e.g.}, HT, BTD, Kronecker tensor decomposition (KTD) \cite{Phan_2012_KTD,Phan_2013_KTD}, PEPS \cite{Schuch_2010_PEPS}, MERA \cite{Li_2018_MERA}, etc. \textbf{Researching these tensor formats may imply potential unrevealed prospects on neural network compression}.
    
    \item Quantization and pruning are usually combined to compress DNNs which can be observed in Table \ref{tab:Joint-compression}. However, in the field of visual recognition, corresponding tasks on RNNs are few. We deem that RNNs have inherent adaptation to sequential visual data such as videos. Therefore, \textbf{we suggest researchers to aim at this direction to work out more compressed RNNs based on quantization and pruning}.
    
    \item As different compression methods have different characteristics, \wang{it is natural to} expect a super-synthetical compression method, which contains the flexibility of NAS, regular architecture of tensor decomposition, efficient computing of quantization in embedded surroundings, and high accuracy of pruning, could be proposed in the future. In other words, \textbf{there are still a lot of works to be done to reach the culmination of network compression}.
    
    \item The miniaturization of DNNs has a great practical significance for embedded devices which may work in manufacturing field, medical facility, aerospace equipment, etc. Hence, \textbf{how to land any of compression methods to various hardware has a broad prospect to study}.
\end{itemize}

\subsubsection{Efficiently Learn Efficient Transferable Models}\quad

Though transfer learning with DNNs has been widely studied due to the lack of labeled data for many visual recognition tasks as stated in subsection \ref{subsec:generalization_TL}, the existing transfer learning models have so far only been optimized for recognition accuracy on target tasks, with the efficiency issue widely disregarded. 

As far as we are aware, there is only one exception, the NASNet work \cite{zoph_2018_CVPR} proposed by Zoph et. al. at CVPR 2018. It designs a search space for NAS that decouples the complexity of an architecture from the depth of a network. The search space permits identifying good architectures on a small dataset (\emph{i.e.}, CIFAR-10) and transferring the learned architecture to image classification datasets/tasks across a range of data and computational scales (including the large ImageNet dataset). The NASNet actually finds the best convolutional layer (or “cell”) on the small dataset and then applys this cell to another dataset by stacking together more copies of this cell, each with their own parameters to composite a convolutional architecture. Therefore, the architecture can be flexible, general and light-weight. The resulting architectures approach or exceed state-of-the-art performance in both CIFAR-10 and ImageNet datasets with less computational demand than human designed architectures \cite{zoph_2018_CVPR}. In greater details, the learned NASNet is 1.2\% better in top-1 accuracy than the best human-invented architectures while having 9 billion fewer FLOPS which is a reduction of 28\% in computational demand from the previous state-of-the-art model. When used with the Faster-RCNN framework, the features learned by NASNet also surpass state-of-the-art at that time by 4.0\% achieving 43.1\% mAP on the COCO dataset. \textbf{The work shows a great example on approaching a more efficient NAS model to learn efficient transferable network architectures.} \emph{Further exploration in this direction will be highly valuable and should be paid more attention.}

\textbf{Besides NAS, adaptive fine-tuning with dynamic routing for transfer learning also has the potential to be able to learn an input-adaptive efficient transferable model.} The very recent work of SpotTune \cite{guo_2019spottune_CVPR} has shown how to learn such an input-adaptive model under a transfer learning setting. However, the dynamic routing component is designed only for improving accuracy but not efficiency. Since such a strategy has been widely used for fast run-time inference as detailed in the former subsection \ref{subsec:network_centric_compression}, it is likely that a unified model may be designed for enhancing both accuracy and efficiency. In any case, \emph{it will be great if efficiency can be taken into account when new DNN-based transfer learning models are designed, so that they can get closer to real applications.}


\ifCLASSOPTIONcaptionsoff
  \newpage
\fi



\bibliographystyle{IEEEtran}
\bibliography{./references.bib}
\end{document}